\documentclass[11pt,a4paper]{article}
\usepackage[hyperref]{acl2020}
\usepackage{times}
\usepackage{latexsym}

\usepackage{mysty}
\usepackage{times}
\usepackage{epsfig}
\usepackage{graphicx}
\usepackage{amsmath}
\usepackage{amssymb}
\usepackage{multirow}
\usepackage{booktabs}
\usepackage{siunitx}
\usepackage{subcaption}
\usepackage{bigstrut}
\usepackage{float}
\usepackage{setspace}

\usepackage{appendix}

\usepackage{tikz}
\usetikzlibrary{decorations.pathmorphing}
\usetikzlibrary{intersections}
\usetikzlibrary{positioning}
\usetikzlibrary{calc}
\usetikzlibrary{fit}
\usepackage[skins]{tcolorbox}

\usepackage{upgreek}
\usepackage{bm}
\makeatletter
\newcommand{\bfgreek}[1]{\bm{\@nameuse{#1}}}
\newcommand{\bfgreekcolor}[2]{\bm{\textcolor{#1}{\@nameuse{#2}}}}
\makeatother

\usepackage[hyperref]{acl2020}
\usepackage{latexsym}

\usepackage{url}
\usepackage{xspace} %

\newif\ifshowcomment
\newcommand{\xxcom}[2]{\ifshowcomment\textcolor{#1}{#2}\fi}
\newcommand{\rtcom}[1]{\xxcom{blue}{rt: #1}} %

\newcommand{\supporapp}[0]{\ifarxiv Appendix\else supplementary material\fi}

\newif\ifarxiv
\arxivtrue

\newcommand{\spicen}{AllSPICE\xspace}

\usepackage{microtype}

\aclfinalcopy %

\title{Analysis of diversity-accuracy tradeoff in image captioning}

\author{Ruotian Luo \\
  TTI-Chicago \\
  {\tt rluo@ttic.edu} \\\And
  Gregory Shakhnarovich \\
  TTI-Chicago \\
  {\tt greg@ttic.edu} \\}

\date{}

\begin{document}
\maketitle

\begin{abstract}

  We investigate the effect of different model
  architectures, training objectives, hyperparameter
  settings and decoding procedures on the diversity of automatically generated
  image captions. Our results show that 1) simple decoding by naive sampling, coupled with
  low temperature is a competitive and fast method to produce diverse and accurate
  caption sets; 2) training with CIDEr-based reward using Reinforcement learning 
  harms the diversity properties of the resulting
  generator, which cannot be mitigated by manipulating decoding parameters. In addition,
  we propose a new metric \spicen for evaluating both accuracy and
  diversity of a set of captions by a single value.

\end{abstract}

\section{Introduction}

People can produce a diverse yet accurate set of captions for a given
image. Sources of diversity include syntactic variations and
paraphrasing, focus on different components of the visual scene, and
focus on different levels of abstraction, e.g., describing scene
composition vs. settings vs. more abstract, non-grounded
circumstances, observed or imagined. In this paper we consider the
goal of endowing automatic image caption generators with the ability to
produce diverse caption sets.

\textbf{Why is caption diversity important}, beyond the fact that ``people can
do it'' (evident in existing data sets with multiple captions per
image, such as COCO)? One caption may not be sufficient to describe
the whole image and there are more than one way to describe an
image. Producing multiple distinct descriptions is required in tasks like image paragraph
generation and dense captioning. Additionally, access to diverse
\emph{and} accurate caption set may enable better production of a
single caption, e.g., when combined with a re-ranking method~\cite{collins2005discriminative, shen2003svm,luo2017comprehension, holtzman2018learning, park2019adversarial}.

\textbf{How is diversity achieved} in automated captioning?
Latent variable models like Conditional GAN~\cite{shetty2017speaking,dai2017towards}, Conditional VAE~\cite{wang2017diverse,jain2017creativity} and Mixture of experts~\cite{wang2016diverse,shen2019mixture}  generate a set of captions by sampling the latent variable.
However, they didn't carefully compare against different sampling methods. For example, the role of sampling temperature in controlling diversity/accuracy tradeoff was overlooked.

Some efforts have been made to explore different sampling methods to generate diverse outputs, like diverse beam search~\cite{vijayakumar2016diverse}, Top-K sampling~\cite{fan2018hierarchical, radford2019language}, nucleus sampling~\cite{Ari2019nucleus} and etc.\cite{batra2012diverse,gimpel2013systematic,li2016mutual,li2015diversity}. We want to study these methods for captioning.

\cite{wang2019describing} has found that existing models have a trade-off between diversity and accuracy. By training a weighted combination of CIDEr reward and cross entropy loss, they can get models with different diversity-accuracy trade-off with different weight factor.

In this paper, we want to take SOTA captioning models, known to achieve good accuracy, and study the diversity accuracy trade-off with different sampling methods and hyperparameters while fixing the model parameters. Especially, we will show if CIDEr-optimized model (high accuracy low diversity) can get a better trade-off than cross-entropy trained model under different sampling settings.
\rtcom{Should we change to quality??}

\textbf{How is diversity measured?} Common metrics for captioning
tasks like BLEU, CIDEr, etc., are aimed at measuring similarity
between a pair of single captions, and can't evaluate diversity. On
the other hand, some metrics have been proposed to capture diversity,
but not accuracy. These include
mBLEU, Distinct unigram,
Distinct bigram, Self-CIDEr~\cite{wang2019describing}. We
propose a new metric \spicen, an extension of SPICE able to measure diversity and accuracy
of a caption set w.r.t. ground truth captions at the same time. Note that, diversity evaluated in \spicen
is defined as the ability of generating different captions for the same image, similar to Self-CIDEr,
as opposed to vocabulary size or word recall~\cite{van2018measuring}, where diversity across
test image set is considered.

In addition to the new metric, our primary contribution is the systematic evaluation of the role of different choices -- training objectives; hyperparameter values; sampling/decoding procedure -- play in the resulting tradeoff between accuracy and the diversity of generated caption sets. This allows us to identify some simple but effective approaches to producing diverse and accurate captions.

 \section{Methods for diverse captioning}
In most of our experiments we work with the attention-based LSTM model~\cite{rennie2017self}
with hidden layer size 512. We consider alternative models (without
attention; larger layer size; and a transformer-based model) in
\supporapp\ and some experiments.

\noindent Learning objectives we consider:

\noindent\textbf{XE}: standard cross entropy loss.

\noindent\textbf{RL}: pretrained with XE, then finetuned on CIDEr reward using REINFORCE~\cite{williams1992simple,rennie2017self}

\noindent\textbf{XE+RL}: pretrained with XE, then finetuned on convex combination of XE and
      CIDEr reward.

Next, we consider different decoding (production) methods for trained
generators.

\noindent\textbf{SP} Naive sampling, with temperature $T$,
      \[p(w_t|I,w_{1\ldots,t})=\frac{\exp\left(s(w_t|I,w_{1,\ldots,t-1})/T\right)}
        {\sum_w\exp\left(s(w_t|I,w_{1,\ldots,t-1})/T\right)}
      \]
      where $s$ is the score (logit) of the model for the
      next word given image $I$. We are
      not aware of any discussion of this approach to 
      diverse captioning in the literature.
      
\noindent\textbf{Top-K} sampling~\cite{fan2018hierarchical, radford2019language} limiting sampling to top $K$ words.

\noindent\textbf{Top-p} (nucleus) sampling~\cite{Ari2019nucleus} limiting
sampling to a set with high probability mass $p$.

\noindent\textbf{BS} Beam search, producing $m$-best list.

\noindent\textbf{DBS} Diverse beam search~\cite{vijayakumar2016diverse} with diversity
parameter $\lambda$.

\noindent Each of these methods can be used with temperature $T$ affecting the word
posterior, as well as its ``native'' hyperparameters ($K$,$p$,$m$,$\lambda$).

\section{Measuring diversity and accuracy}
We will use use the following metrics for diverse captioning:

\noindent\textbf{mBLEU} Mean of BLEU scores computed between each caption
in the set against the rest. Lower=more diversity.

\noindent\textbf{Div-1, Div-2}: ratio of number of unique uni/bigrams 
    in generated captions to number of words in caption set.
    Higher=more diverse.
    
\noindent\textbf{Self-CIDEr}\cite{wang2019describing}: a diversity metric derived from latent 
    semantic analysis and %
    CIDEr similarity. Higher=more diverse.

\noindent\textbf{Vocabulary Size}: number of unique words used in all
generated captions.

\noindent\textbf{\% Novel Sentences}: percentage of generated captions
not seen in the training set.

\noindent\textbf{Oracle/Average} scores: taking the
maximum/mean of each relevant metric over 
all the candidates (of one image).

We propose a new metric \spicen. It builds upon the SPICE~\cite{anderson2016spice} designed for
evaluating a single caption. To compute SPICE, a \emph{scene graph} is
constructed for the evaluated caption, with vertices corresponding to
objects, attributes, and relations; edges reflect possession of attributes or relations between objects. Another scene graph is constructed from the reference
caption; with multiple reference captions, the reference scene graph
is the union of the graphs for individual captions, combining
synonymous vertices. The SPICE metric is the F-score
on matching vertices and edges between the two graphs.

\spicen builds a single graph for the
\emph{generated caption set}, in the same way that SPICE treats
reference caption sets. Note: repetitions across captions
in the set won't change the score, due to merging of synonymous
vertices. Adding a caption that captures part of the reference not
 captured by previous captions in the set may improve the
 score (by increasing recall). Finally, wrong content in any caption in the set will harm the
score (by reducing precision). 
This is in contrast to ``oracle score'', where only one caption in the set
needs to be good, and adding wrong captions does not affect the
metric, making it an imperfect measure of \emph{set} quality.

Thus, higher \spicen requires the samples to be semantically diverse
\emph{and} correct; diversity without accuracy is penalized. This
is in contrast to other metrics (mBLEU, Self-CIDEr),
that separate diversity
considerations from those of accuracy.\ifaclfinal \footnote{AllSPICE is released at \url{https://github.com/ruotianluo/coco-caption}}\fi

\section{Experiments}
The aim of our experiments is to explore how different
methods (varying training procedures, decoding
procedures, and hyper-parameter settings) navigate the
diversity/accuracy tradeoff. We use data from COCO
captions~\cite{lin2014microsoft}, with splits
from~\cite{karpathy2015deep}, reporting results on the test split.
(For similar results on Flickr30k~\cite{young-etal-2014-image} see \supporapp.)
Unless stated otherwise, caption sets contain five generated
captions for each image. \ifaclfinal \footnote{The experiment related code is released at in \url{https://github.com/ruotianluo/self-critical.pytorch/tree/master/projects/Diversity}}\fi %

\begin{figure}[th]
\begin{minipage}[t]{.45\linewidth}
\includegraphics[height=2.5cm]{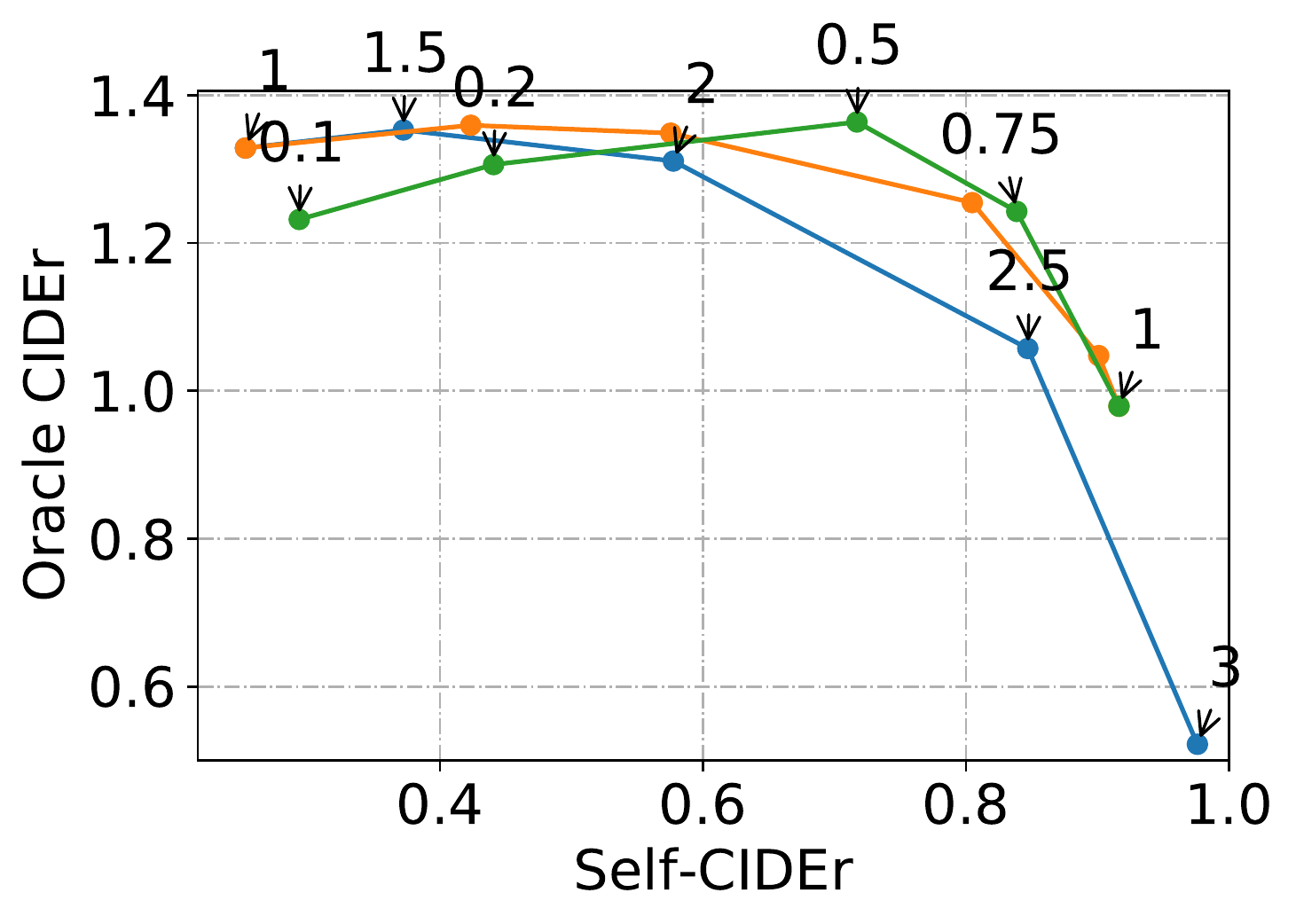}\\
\end{minipage}%
\hspace{.25em}
\begin{minipage}[t]{.45\linewidth}
\includegraphics[height=2.4cm]{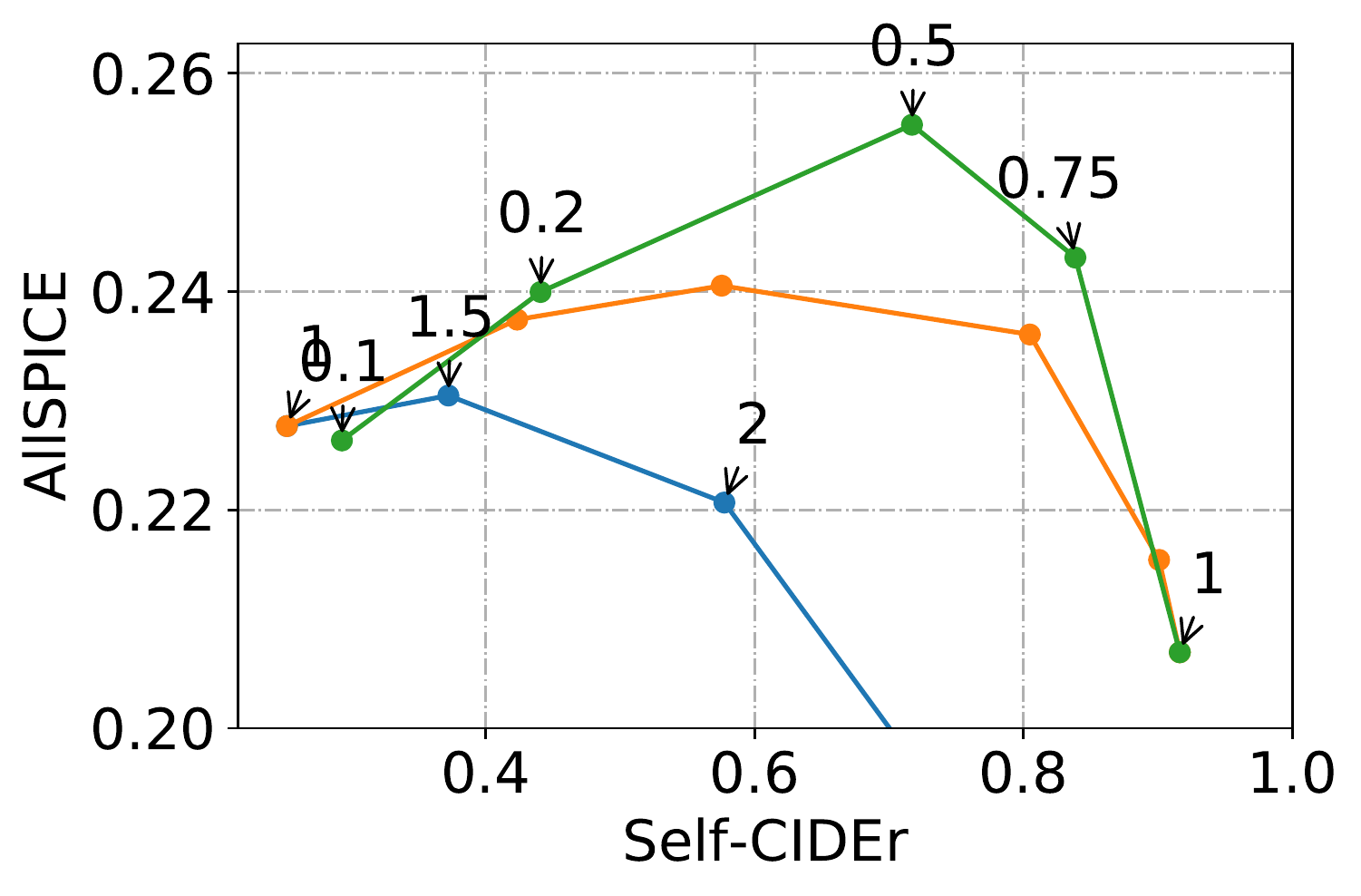}\\
\end{minipage}
\vspace{-2em}\caption{Additional views of diversity/accuracy tradeoff: Oracle CIDEr and \spicen vs. self-CIDEr. See legend for Figure~\ref{fig:1}.}
\label{fig:2}
\end{figure}

\noindent\textbf{Naive sampling} \cite{wang2019describing} note that
with temperature 1, XE-trained model will have high diversity but low
accuracy, and RL-trained model the opposite. Thus they propose to use
a weighted combination of RL-objective and XE-loss to achieve better
trade-off. However, Figure \ref{fig:1} shows that a simple alternative
for trading diversity for accuracy is to modulate the sampling
temperature. Although RL-trained model is not diverse when $T$=1, it
generates more diverse captions with higher
temperature. XE-trained model can generate \emph{more accurate} captions
with lower temperature.
In Figure \ref{fig:2}, we show that naive sampling with $T$=0.5 performs better than other settings on oracle CIDEr and \spicen, as well.

We also see, comparing the two panels of Figure~\ref{fig:2} to each other and to Figure~\ref{fig:1} that measuring \spicen allows significant distinction between settings that would appear very similar under oracle CIDEr. This suggests that \spicen is complementary to other metrics, and reflects a different aspect of accuracy/diversity tradeoff.

\begin{figure}[t]
\centering
\includegraphics[width=\linewidth]{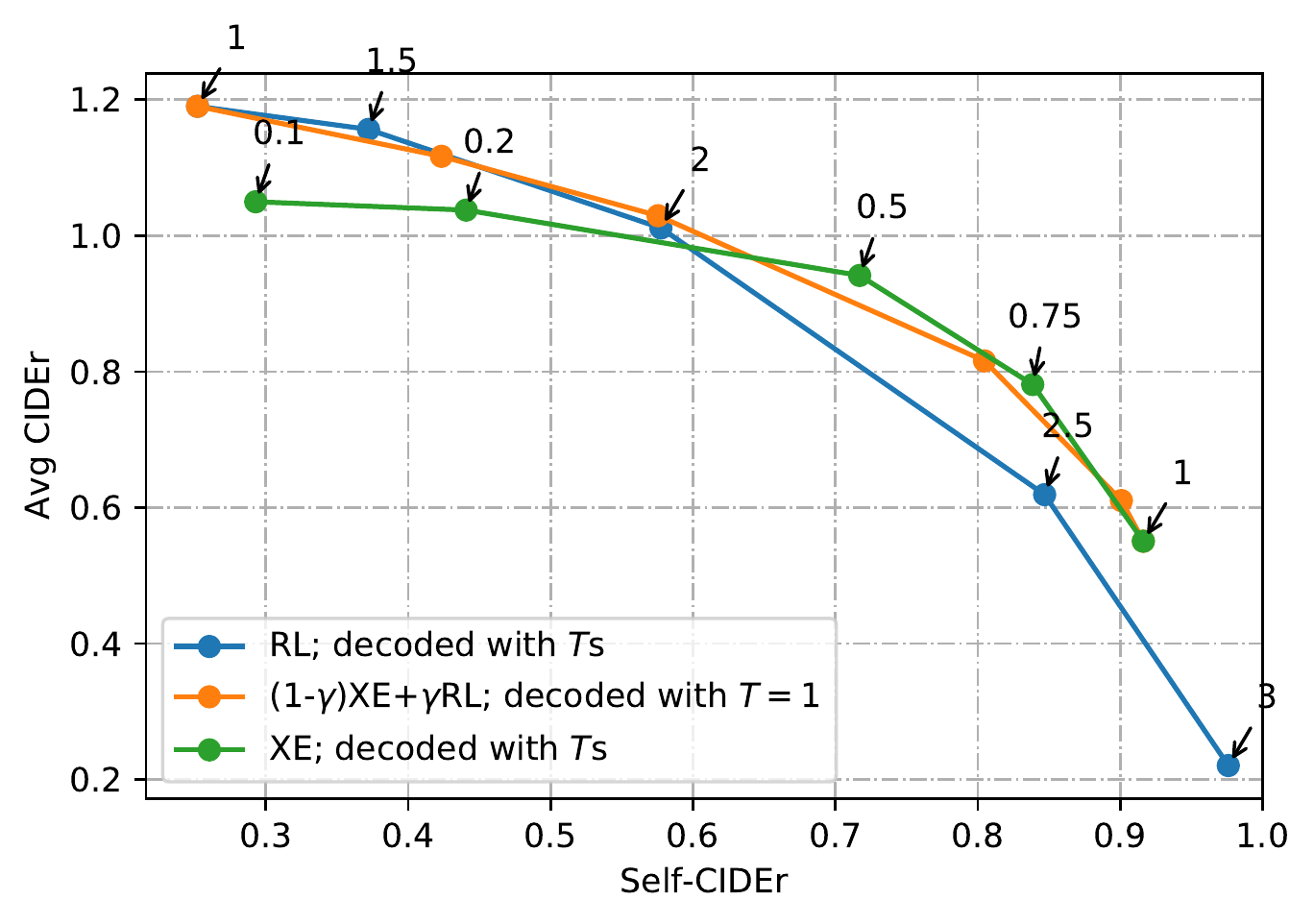}
\caption{Diversity-accuracy tradeoff. Blue: RL, decoding with different
  temperatures. Orange: XE+RL, trained with different
  CIDEr-MLE weights $\gamma$, decoding with $T=1$. Green: XE, decoding
  with different temperatures.\vspace{-1em}}%
\label{fig:1}
\end{figure}

{\renewcommand{\arraystretch}{.8}
\begin{table}[tb!]\small
  \centering
  \scalebox{0.8}{
  \setlength\tabcolsep{3pt}
    \begin{tabular}{lrrrr}
          & \multicolumn{1}{c}{avg CIDEr} & \multicolumn{1}{c}{oracle CIDEr} & \multicolumn{1}{c}{\spicen} & \multicolumn{1}{c}{Self-CIDEr} \\
    \midrule
    DBS $\lambda$=0.3, T=1 & 0.919 & 1.413 & \textbf{0.271} & \textbf{0.754} \\
    BS T=0.75 & \textbf{1.073} & \textbf{1.444} & 0.261 & 0.588 \\
    Top-K K=3 T=0.75 & 0.921 & 1.365 & 0.258 & 0.736 \\
    Top-p p=0.8 T=0.75 & 0.929 & 1.366 & 0.257 & 0.744 \\
    SP T=0.5 & 0.941 & 1.364 & 0.255 & 0.717 \\
    \bottomrule
    \end{tabular}
}
  \caption{Performnace for best hyperparamters for each method (XE-trained model).}
  \label{tab:methods_sp_5}%
\end{table}}

\noindent\textbf{Biased sampling}
 Top-K sampling~\cite{fan2018hierarchical, radford2019language} and Top-p (nucleus) sampling\cite{Ari2019nucleus} are used to eliminate low-probability words by only sampling among top probability choices. This is intended to prevent occurrence of ``bad'' words, which, once emitted, poison subsequent generation.

\begin{figure}[tb!]
\begin{minipage}[t]{.45\linewidth}
\includegraphics[height=2.4cm]{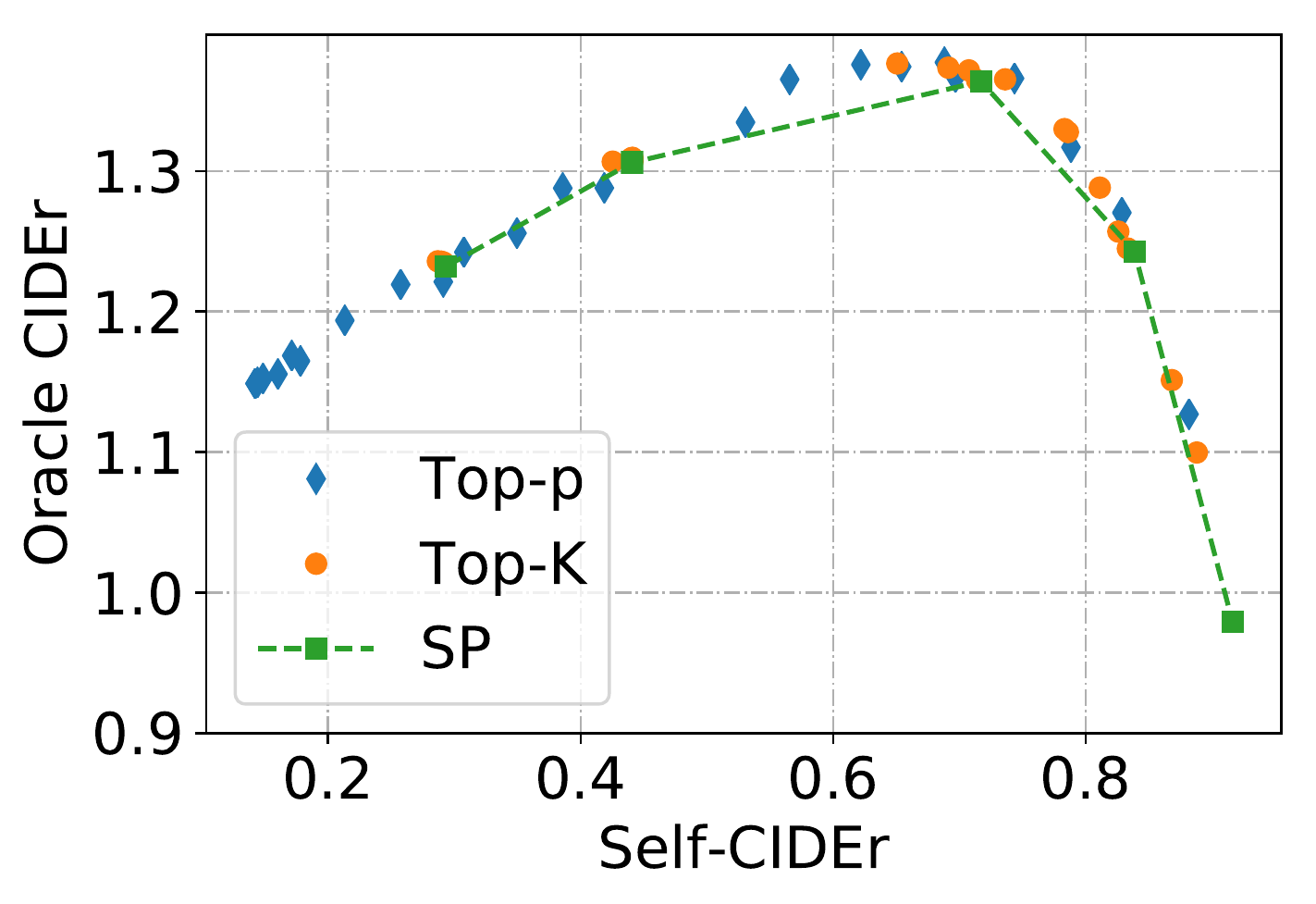}\\
\end{minipage}%
\hspace{.25em}
\begin{minipage}[t]{.45\linewidth}
\includegraphics[height=2.4cm]{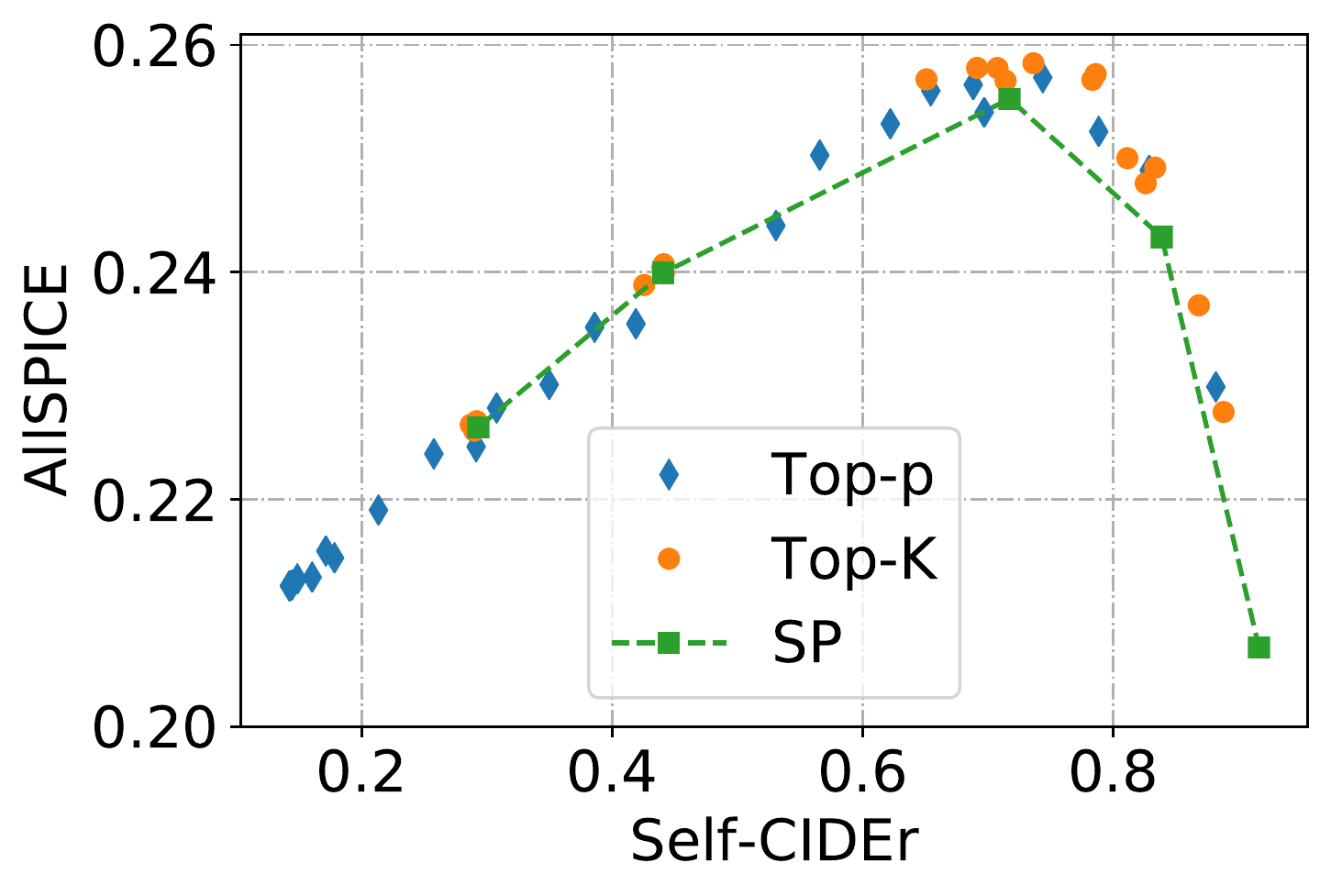}\\
\end{minipage}
\vspace{-1.5em}\caption{Oracle CIDEr and \spicen of Top-K (orange) and Top-p (blue)
  sampling, compared to naive sampling with varying $T$
  (green), for XE-trained model.}%
\label{fig:3}
\end{figure}

In Figure \ref{fig:3}, we explore these methods for XE-trained models,
with  different thresholds and temperatures, compared against SP/temperature.
We see that they can slightly outperform SP.
In \supporapp, we show that Top-K sampling with RL-trained model gets better results than SP or Top-p, however, the results are still worse than XE-trained models.

\noindent\textbf{Beam search}: we include in the \supporapp\ 
the detailed curves comparing
results of (regular and diverse) beam search with varying temperature on XE and RL
models. Briefly, we observe that as with sampling, RL trained models
exhibit worse diversity/accuracy tradeoff when BS is used for decoding.

In summary, in \emph{all settings}, the RL-trained model performs worse than XE-trained models.
suggesting that RL-objective is detrimental to diversity.
In the remainder, we only discuss XE-trained models. 

\noindent\textbf{Comparing across methods (sample size 5)}
Table \ref{tab:methods_sp_5} shows the performance of each methods under its best performing hyperparameter (under \spicen). Diverse beam search is the best algorithm with high \spicen and Self-CIDEr, indicating both semantic and syntactic diversity.

Beam search performs best on oracle CIDEr and average CIDEr, and it performs well on \spicen too. However although all the generated captions are accurate, the syntactic diversity is missing, shown by Self-CIDEr. Qualitative results are shown in supplementary.

Sampling methods (SP, Top-K, Top-p) are much faster than beam
search, and competitive in performance. This suggests them as a compelling alternative to recent methods for speeding up diverse caption generation, like~\cite{deshpande2018diverse}.

\begin{figure}[t]
\centering
\includegraphics[width=\linewidth]{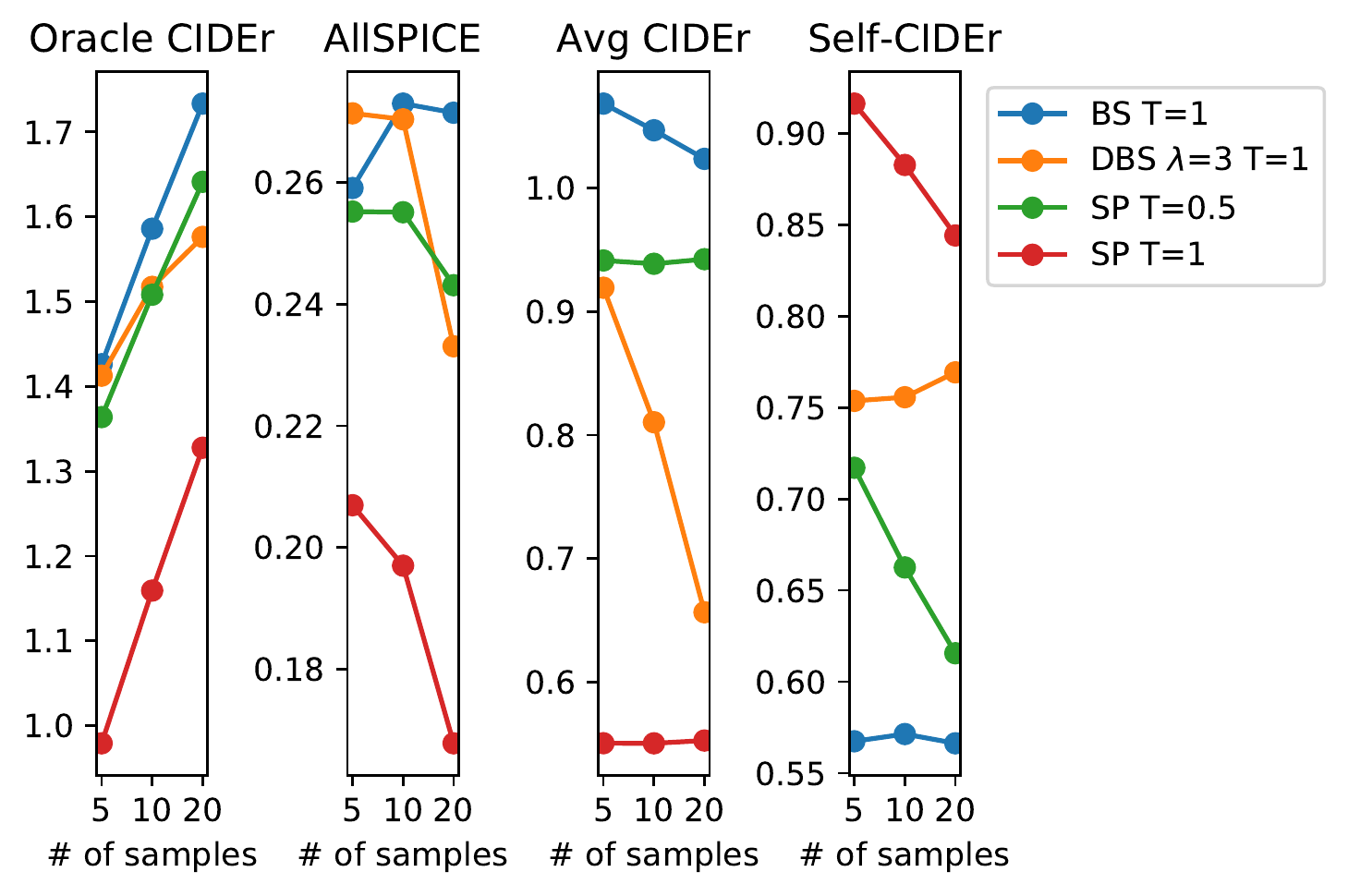}
\caption{Effect of sample size on different metrics, XE-trained model. Colors indicate decoding strategies.}
\label{fig:samplen}
\end{figure}

\noindent\textbf{Different sample size} (see
Figure~\ref{fig:samplen}): \emph{Oracle CIDEr} tends to increase with sample size, 
as more captions mean more chances to fit the reference. 

\noindent\emph{\spicen} drops with more samples, because additional captions are more likely to hurt (say something wrong) than help (add something correct not yet said). BS, which explores the caption space more ``cautiously'' than other methods, is initially resilient to this effect, but with enough samples its \spicen drops as well.

\noindent\emph{Average CIDEr} Sampling methods' average scores are largely invariant to sample size.  BS and especially DBS suffer a lot with more samples, because diversity constraints and the properties of the beam search force the additional captions to be lower quality, hurting precision without improving recall.

\noindent\emph{Self-CIDEr} The relative order in Self-CIDEr is the same across different sample size; but SP/$T$=1 is closer to DBS and SP/$T$=0.5 to BS in Self-CIDEr. Because of the definition of self-CIDEr, comparison across sample sizes is not very meaningful.

{\renewcommand{\arraystretch}{.8}
\begin{table}[!tb]\small
  \centering
  \scalebox{0.8}{
  \setlength\tabcolsep{3pt}
    \begin{tabular}{lrrrrrrr}
          & \multicolumn{1}{c}{\multirow{2}[1]{*}{METEOR}} & \multicolumn{1}{c}{\multirow{2}[1]{*}{SPICE}} & \multicolumn{1}{c}{\multirow{2}[1]{*}{Div-1}} & \multicolumn{1}{c}{\multirow{2}[1]{*}{Div-2}} & \multicolumn{1}{c}{\multirow{2}[1]{*}{mBleu-4}} & \multicolumn{1}{c}{Vocab-} & \multicolumn{1}{c}{\%Novel } \\
          &       &       &       &       &       & \multicolumn{1}{c}{ulary} & \multicolumn{1}{c}{Sent} \\
    \hline
    Base & 0.265 & 0.186 & 0.31  & 0.44  & 0.68  & 1460  & 55.2 \\
    Adv & 0.236 & 0.166 & 0.41  & 0.55  & 0.51  & 2671  & 79.8 \\
    \hline
    T=0.33 & 0.266 & 0.187 & 0.31  & 0.44 & 0.65 & 1219 & 58.3 \\
    T=0.5 & 0.260 & 0.183 & 0.37  & 0.55  & 0.47  & 1683  & 75.6 \\
    T=0.6 & 0.259 & 0.181 & 0.41  & 0.60  & 0.37  & 2093  & 83.7 \\
    T=0.7 & 0.250 & 0.174 & 0.45  & 0.66  & 0.28  & 2573  & 89.9 \\
    \textcolor{green!50!black}{T=0.8} & 0.240 & 0.166 & 0.49  & 0.71  & 0.21  & 3206  & 94.4 \\
    T=0.9 & 0.228 & 0.157 & 0.54  & 0.77  & 0.14  & 3969  & 97.1 \\
    T=1   & 0.214 & 0.144 & 0.59  & 0.81  & 0.08  & 4875  & 98.7 \\
    \bottomrule
    \end{tabular}%
}
  \caption{Base and Adv are from \cite{shetty2017speaking}. The other models are trained by us. All the methods use naive sampling decoding. }
  \label{tab:gan}%
\end{table}}

\noindent\textbf{Comparison with adversarial loss}
In \cite{shetty2017speaking}, the authors get more diverse but less
accurate captions with adversarial loss. Table~\ref{tab:gan} shows that by tuning sampling temperature, a XE-trained model can generate more diverse \emph{and} more accurate captions than adversarially trained model. Following~\cite{shetty2017speaking}, METEOR and SPICE are computed by measuring the score of the highest likelihood sample among 5 generated captions.
For fair comparison, we use a similar non-attention LSTM model. When temperature is 0.8, the output generations outperforms the adversarial method (Adv). While \cite{shetty2017speaking} also compare their model to a base model using sampling method (Base), they get less diverse captions because they use temperature $\frac{1}{3}$ (our model with T=0.33 is performing similar to ``Base'').

 \section{Conclusions}

Our results provide
both a practical guidance and a better understanding of the
diversity/accuracy tradeoff in image captioning.
We show that caption decoding methods may affect
this tradeoff more significantly than the choice of
training objectives or model. In particular, simple naive sampling, coupled with suitably low
temperature, is a competitive method with respect to speed and
diversity/accuracy tradeoff. Diverse beam search exhibits the best
tradeoff, but is also the slowest.
Among training objectives, using CIDEr-based
reward reduces diversity in a way that is not mitigated by
manipulating decoding parameters. Finally, we introduce a
metric \spicen for directly evaluating tradeoff between accuracy and semantic diversity.

\bibliographystyle{acl_natbib}
\bibliography{acl2020}

\begin{thebibliography}{37}
\expandafter\ifx\csname natexlab\endcsname\relax\def\natexlab#1{#1}\fi

\bibitem[{Anderson et~al.(2016)Anderson, Fernando, Johnson, and
  Gould}]{anderson2016spice}
Peter Anderson, Basura Fernando, Mark Johnson, and Stephen Gould. 2016.
\newblock Spice: Semantic propositional image caption evaluation.
\newblock In \emph{European Conference on Computer Vision}, pages 382--398.
  Springer.

\bibitem[{Anderson et~al.(2018)Anderson, He, Buehler, Teney, Johnson, Gould,
  and Zhang}]{anderson2018bottom}
Peter Anderson, Xiaodong He, Chris Buehler, Damien Teney, Mark Johnson, Stephen
  Gould, and Lei Zhang. 2018.
\newblock Bottom-up and top-down attention for image captioning and visual
  question answering.
\newblock In \emph{Proceedings of the IEEE Conference on Computer Vision and
  Pattern Recognition}, pages 6077--6086.

\bibitem[{Batra et~al.(2012)Batra, Yadollahpour, Guzman-Rivera, and
  Shakhnarovich}]{batra2012diverse}
Dhruv Batra, Payman Yadollahpour, Abner Guzman-Rivera, and Gregory
  Shakhnarovich. 2012.
\newblock Diverse m-best solutions in markov random fields.
\newblock In \emph{European Conference on Computer Vision}, pages 1--16.
  Springer.

\bibitem[{Collins and Koo(2005)}]{collins2005discriminative}
Michael Collins and Terry Koo. 2005.
\newblock Discriminative reranking for natural language parsing.
\newblock \emph{Computational Linguistics}, 31(1):25--70.

\bibitem[{Dai et~al.(2017)Dai, Fidler, Urtasun, and Lin}]{dai2017towards}
Bo~Dai, Sanja Fidler, Raquel Urtasun, and Dahua Lin. 2017.
\newblock Towards diverse and natural image descriptions via a conditional gan.
\newblock In \emph{Proceedings of the IEEE International Conference on Computer
  Vision}, pages 2970--2979.

\bibitem[{Deshpande et~al.(2018)Deshpande, Aneja, Wang, Schwing, and
  Forsyth}]{deshpande2018diverse}
Aditya Deshpande, Jyoti Aneja, Liwei Wang, Alexander Schwing, and David~A
  Forsyth. 2018.
\newblock Diverse and controllable image captioning with part-of-speech
  guidance.
\newblock \emph{arXiv preprint arXiv:1805.12589}.

\bibitem[{Fan et~al.(2018)Fan, Lewis, and Dauphin}]{fan2018hierarchical}
Angela Fan, Mike Lewis, and Yann Dauphin. 2018.
\newblock Hierarchical neural story generation.
\newblock \emph{arXiv preprint arXiv:1805.04833}.

\bibitem[{Gimpel et~al.(2013)Gimpel, Batra, Dyer, and
  Shakhnarovich}]{gimpel2013systematic}
Kevin Gimpel, Dhruv Batra, Chris Dyer, and Gregory Shakhnarovich. 2013.
\newblock A systematic exploration of diversity in machine translation.
\newblock In \emph{Proceedings of the 2013 Conference on Empirical Methods in
  Natural Language Processing}, pages 1100--1111.

\bibitem[{He et~al.(2016)He, Zhang, Ren, and Sun}]{he2016deep}
Kaiming He, Xiangyu Zhang, Shaoqing Ren, and Jian Sun. 2016.
\newblock Deep residual learning for image recognition.
\newblock In \emph{Proceedings of the IEEE conference on computer vision and
  pattern recognition}, pages 770--778.

\bibitem[{Holtzman et~al.(2018)Holtzman, Buys, Forbes, Bosselut, Golub, and
  Choi}]{holtzman2018learning}
Ari Holtzman, Jan Buys, Maxwell Forbes, Antoine Bosselut, David Golub, and
  Yejin Choi. 2018.
\newblock Learning to write with cooperative discriminators.
\newblock \emph{arXiv preprint arXiv:1805.06087}.

\bibitem[{Holtzman et~al.(2019)Holtzman, Buys, Forbes, and
  Choi}]{Ari2019nucleus}
Ari Holtzman, Jan Buys, Maxwell Forbes, and Yejin Choi. 2019.
\newblock \href {http://arxiv.org/abs/arXiv:1904.09751} {The curious case of
  neural text degeneration}.

\bibitem[{Jain et~al.(2017)Jain, Zhang, and Schwing}]{jain2017creativity}
Unnat Jain, Ziyu Zhang, and Alexander~G Schwing. 2017.
\newblock Creativity: Generating diverse questions using variational
  autoencoders.
\newblock In \emph{Proceedings of the IEEE Conference on Computer Vision and
  Pattern Recognition}, pages 6485--6494.

\bibitem[{Karpathy and Fei-Fei(2015)}]{karpathy2015deep}
Andrej Karpathy and Li~Fei-Fei. 2015.
\newblock Deep visual-semantic alignments for generating image descriptions.
\newblock In \emph{Proceedings of the IEEE conference on computer vision and
  pattern recognition}, pages 3128--3137.

\bibitem[{Kim et~al.(2019)Kim, Rush, Yu, Kuncoro, Dyer, and
  Melis}]{kim2019unsupervised}
Yoon Kim, Alexander~M Rush, Lei Yu, Adhiguna Kuncoro, Chris Dyer, and G{\'a}bor
  Melis. 2019.
\newblock Unsupervised recurrent neural network grammars.
\newblock \emph{arXiv preprint arXiv:1904.03746}.

\bibitem[{Kingma and Ba(2014)}]{kingma2014adam}
Diederik~P Kingma and Jimmy Ba. 2014.
\newblock Adam: A method for stochastic optimization.
\newblock \emph{arXiv preprint arXiv:1412.6980}.

\bibitem[{Krishna et~al.(2017)Krishna, Zhu, Groth, Johnson, Hata, Kravitz,
  Chen, Kalantidis, Li, Shamma et~al.}]{krishna2017visual}
Ranjay Krishna, Yuke Zhu, Oliver Groth, Justin Johnson, Kenji Hata, Joshua
  Kravitz, Stephanie Chen, Yannis Kalantidis, Li-Jia Li, David~A Shamma, et~al.
  2017.
\newblock Visual genome: Connecting language and vision using crowdsourced
  dense image annotations.
\newblock \emph{International Journal of Computer Vision}, 123(1):32--73.

\bibitem[{Li et~al.(2015)Li, Galley, Brockett, Gao, and
  Dolan}]{li2015diversity}
Jiwei Li, Michel Galley, Chris Brockett, Jianfeng Gao, and Bill Dolan. 2015.
\newblock A diversity-promoting objective function for neural conversation
  models.
\newblock \emph{arXiv preprint arXiv:1510.03055}.

\bibitem[{Li and Jurafsky(2016)}]{li2016mutual}
Jiwei Li and Dan Jurafsky. 2016.
\newblock Mutual information and diverse decoding improve neural machine
  translation.
\newblock \emph{arXiv preprint arXiv:1601.00372}.

\bibitem[{Lin et~al.(2014)Lin, Maire, Belongie, Hays, Perona, Ramanan,
  Doll{\'a}r, and Zitnick}]{lin2014microsoft}
Tsung-Yi Lin, Michael Maire, Serge Belongie, James Hays, Pietro Perona, Deva
  Ramanan, Piotr Doll{\'a}r, and C~Lawrence Zitnick. 2014.
\newblock Microsoft coco: Common objects in context.
\newblock In \emph{European conference on computer vision}, pages 740--755.
  Springer.

\bibitem[{Luo and Shakhnarovich(2017)}]{luo2017comprehension}
Ruotian Luo and Gregory Shakhnarovich. 2017.
\newblock Comprehension-guided referring expressions.
\newblock In \emph{Proceedings of the IEEE Conference on Computer Vision and
  Pattern Recognition}, pages 7102--7111.

\bibitem[{van Miltenburg et~al.(2018)van Miltenburg, Elliott, and
  Vossen}]{van2018measuring}
Emiel van Miltenburg, Desmond Elliott, and Piek Vossen. 2018.
\newblock Measuring the diversity of automatic image descriptions.
\newblock In \emph{Proceedings of the 27th International Conference on
  Computational Linguistics}, pages 1730--1741.

\bibitem[{Mnih and Rezende(2016)}]{mnih2016variational}
Andriy Mnih and Danilo~J Rezende. 2016.
\newblock Variational inference for monte carlo objectives.
\newblock \emph{arXiv preprint arXiv:1602.06725}.

\bibitem[{Park et~al.(2019)Park, Rohrbach, Darrell, and
  Rohrbach}]{park2019adversarial}
Jae~Sung Park, Marcus Rohrbach, Trevor Darrell, and Anna Rohrbach. 2019.
\newblock Adversarial inference for multi-sentence video description.
\newblock In \emph{Proceedings of the IEEE Conference on Computer Vision and
  Pattern Recognition}, pages 6598--6608.

\bibitem[{Radford et~al.(2019)Radford, Wu, Child, Luan, Amodei, and
  Sutskever}]{radford2019language}
Alec Radford, Jeffrey Wu, Rewon Child, David Luan, Dario Amodei, and Ilya
  Sutskever. 2019.
\newblock Language models are unsupervised multitask learners.
\newblock \emph{OpenAI Blog}, 1:8.

\bibitem[{Ren et~al.(2015)Ren, He, Girshick, and Sun}]{ren2015faster}
Shaoqing Ren, Kaiming He, Ross Girshick, and Jian Sun. 2015.
\newblock Faster r-cnn: Towards real-time object detection with region proposal
  networks.
\newblock In \emph{Advances in neural information processing systems}, pages
  91--99.

\bibitem[{Rennie et~al.(2017)Rennie, Marcheret, Mroueh, Ross, and
  Goel}]{rennie2017self}
Steven~J Rennie, Etienne Marcheret, Youssef Mroueh, Jerret Ross, and Vaibhava
  Goel. 2017.
\newblock Self-critical sequence training for image captioning.
\newblock In \emph{Proceedings of the IEEE Conference on Computer Vision and
  Pattern Recognition}, pages 7008--7024.

\bibitem[{Sharma et~al.(2018)Sharma, Ding, Goodman, and
  Soricut}]{sharma2018conceptual}
Piyush Sharma, Nan Ding, Sebastian Goodman, and Radu Soricut. 2018.
\newblock Conceptual captions: A cleaned, hypernymed, image alt-text dataset
  for automatic image captioning.
\newblock In \emph{Proceedings of the 56th Annual Meeting of the Association
  for Computational Linguistics (Volume 1: Long Papers)}, pages 2556--2565.

\bibitem[{Shen and Joshi(2003)}]{shen2003svm}
Libin Shen and Aravind~K Joshi. 2003.
\newblock An svm-based voting algorithm with application to parse reranking.
\newblock In \emph{Proceedings of the seventh conference on Natural language
  learning at HLT-NAACL 2003}.

\bibitem[{Shen et~al.(2019)Shen, Ott, Auli, and Ranzato}]{shen2019mixture}
Tianxiao Shen, Myle Ott, Michael Auli, and Marc'Aurelio Ranzato. 2019.
\newblock Mixture models for diverse machine translation: Tricks of the trade.
\newblock \emph{arXiv preprint arXiv:1902.07816}.

\bibitem[{Shetty et~al.(2017)Shetty, Rohrbach, Anne~Hendricks, Fritz, and
  Schiele}]{shetty2017speaking}
Rakshith Shetty, Marcus Rohrbach, Lisa Anne~Hendricks, Mario Fritz, and Bernt
  Schiele. 2017.
\newblock Speaking the same language: Matching machine to human captions by
  adversarial training.
\newblock In \emph{Proceedings of the IEEE International Conference on Computer
  Vision}, pages 4135--4144.

\bibitem[{Vaswani et~al.(2017)Vaswani, Shazeer, Parmar, Uszkoreit, Jones,
  Gomez, Kaiser, and Polosukhin}]{vaswani2017attention}
Ashish Vaswani, Noam Shazeer, Niki Parmar, Jakob Uszkoreit, Llion Jones,
  Aidan~N Gomez, {\L}ukasz Kaiser, and Illia Polosukhin. 2017.
\newblock Attention is all you need.
\newblock In \emph{Advances in neural information processing systems}, pages
  5998--6008.

\bibitem[{Vijayakumar et~al.(2016)Vijayakumar, Cogswell, Selvaraju, Sun, Lee,
  Crandall, and Batra}]{vijayakumar2016diverse}
Ashwin~K Vijayakumar, Michael Cogswell, Ramprasath~R Selvaraju, Qing Sun,
  Stefan Lee, David Crandall, and Dhruv Batra. 2016.
\newblock Diverse beam search: Decoding diverse solutions from neural sequence
  models.
\newblock \emph{arXiv preprint arXiv:1610.02424}.

\bibitem[{Wang et~al.(2017)Wang, Schwing, and Lazebnik}]{wang2017diverse}
Liwei Wang, Alexander Schwing, and Svetlana Lazebnik. 2017.
\newblock Diverse and accurate image description using a variational
  auto-encoder with an additive gaussian encoding space.
\newblock In \emph{Advances in Neural Information Processing Systems}, pages
  5756--5766.

\bibitem[{Wang and Chan(2019)}]{wang2019describing}
Qingzhong Wang and Antoni~B Chan. 2019.
\newblock Describing like humans: on diversity in image captioning.
\newblock \emph{arXiv preprint arXiv:1903.12020}.

\bibitem[{Wang et~al.(2016)Wang, Wu, Lu, Xiao, Li, Zhang, and
  Zhuang}]{wang2016diverse}
Zhuhao Wang, Fei Wu, Weiming Lu, Jun Xiao, Xi~Li, Zitong Zhang, and Yueting
  Zhuang. 2016.
\newblock Diverse image captioning via grouptalk.
\newblock In \emph{IJCAI}, pages 2957--2964.

\bibitem[{Williams(1992)}]{williams1992simple}
Ronald~J Williams. 1992.
\newblock Simple statistical gradient-following algorithms for connectionist
  reinforcement learning.
\newblock \emph{Machine learning}, 8(3-4):229--256.

\bibitem[{Young et~al.(2014)Young, Lai, Hodosh, and
  Hockenmaier}]{young-etal-2014-image}
Peter Young, Alice Lai, Micah Hodosh, and Julia Hockenmaier. 2014.
\newblock \href {https://doi.org/10.1162/tacl_a_00166} {From image descriptions
  to visual denotations: New similarity metrics for semantic inference over
  event descriptions}.
\newblock \emph{Transactions of the Association for Computational Linguistics},
  2:67--78.

\end{thebibliography}

\ifarxiv
\newpage
\appendix
\appendixpage

\section{Training details}
All of our captioning models are trained according to the following scheme. The XE-trained captioning models are trained for 30 epochs (transformer 15 epochs) with Adam\cite{kingma2014adam}. RL-trained models are finetuned from XE-trained model, optimizing CIDEr score with REINFORCE~\cite{williams1992simple}. The baseline used REINFORCE algorithm follows \cite{mnih2016variational, kim2019unsupervised}. This baseline performs better than self critical baseline in \cite{rennie2017self}. The batch size is chosen to be 250 for LSTM-based model, and for transformer the batch size is 100. The learning rate is initialized to be 5e-4 and decay by a factor 0.8 for every three epochs. During reinforcement learning phase, the learning rate is fixed to 4e-5.

\noindent\textbf{The visual features} 
For non-attention LSTM captioning model, we used the input of the last fully connected layer of a pretrained Resnet-101\cite{he2016deep} as image feature.

For attention-based LSTM and transformer, the spatial features are extracted from output of a Faster
R-CNN\cite{anderson2018bottom,ren2015faster} with ResNet-101\cite{he2016deep}, trained by object and attribute annotations from Visual Genome \cite{krishna2017visual}.The output feature has shape Kx2048 where K is the number of detected objects in the image. Both the fully-connected features and spatial features are pre-extracted, and no finetuning is applied on image encoders.

\section{RL-trained model with TopK and Nucleus sampling}

\begin{figure}[th!]
\centering
\includegraphics[height=4cm]{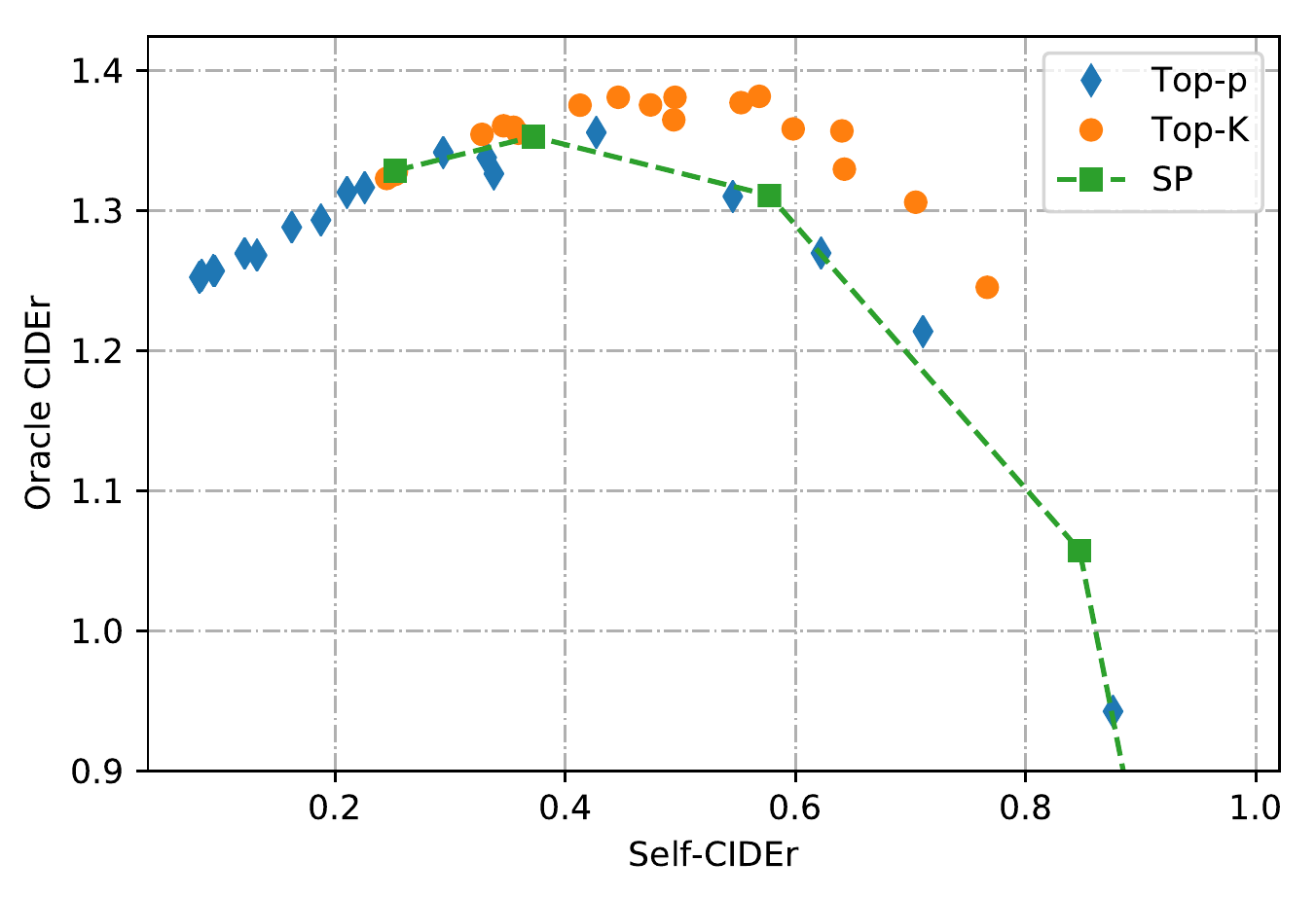}\\
\includegraphics[height=4cm]{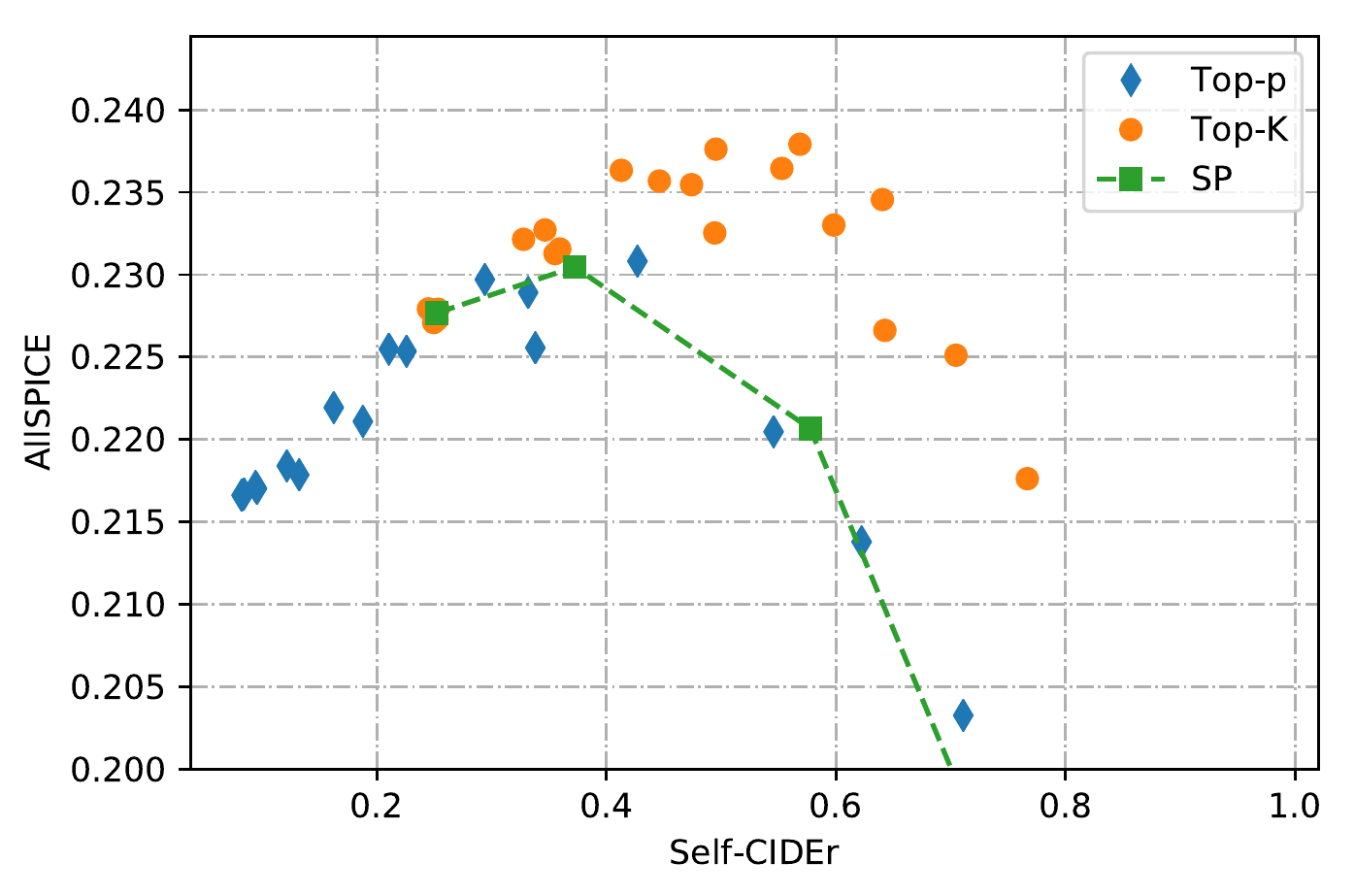}\\
\caption{Oracle CIDEr and AllSPICE of Top-K (orange) and Top-p (blue)
  sampling, compared to naive sampling with varying $T$
  (green), for RL-trained model.}
\label{fig:1}
\end{figure}

For RL-trained model, Top-K sampling can outperform SP and nucleus sampling: captions are more accurate with similar level of diversity, shown in Figure \ref{fig:1}. Our intuition is CIDEr optimization makes the word posterior very peaky and the tail distribution is noisy. With large temperature, the sampling would fail because the selection will fall into the bad tail. Top-K can successfully eliminate this phenomenon and focus on correct things. Unfortunately, nucleus sampling is not performing as satisfying in captioning task.

\section{Beam search}

\begin{figure}[th]
\centering
\includegraphics[height=4cm]{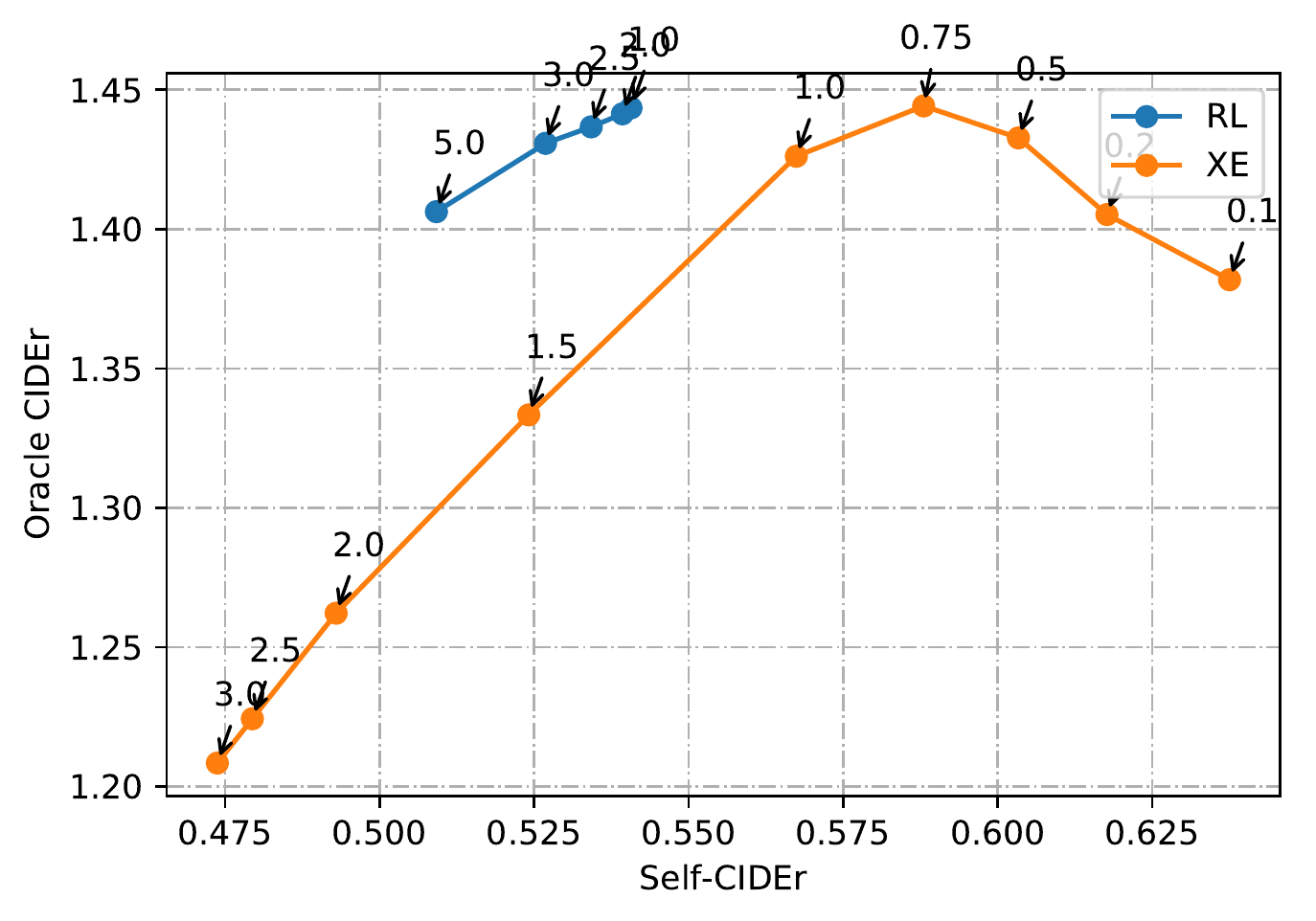}\\
\includegraphics[height=4cm]{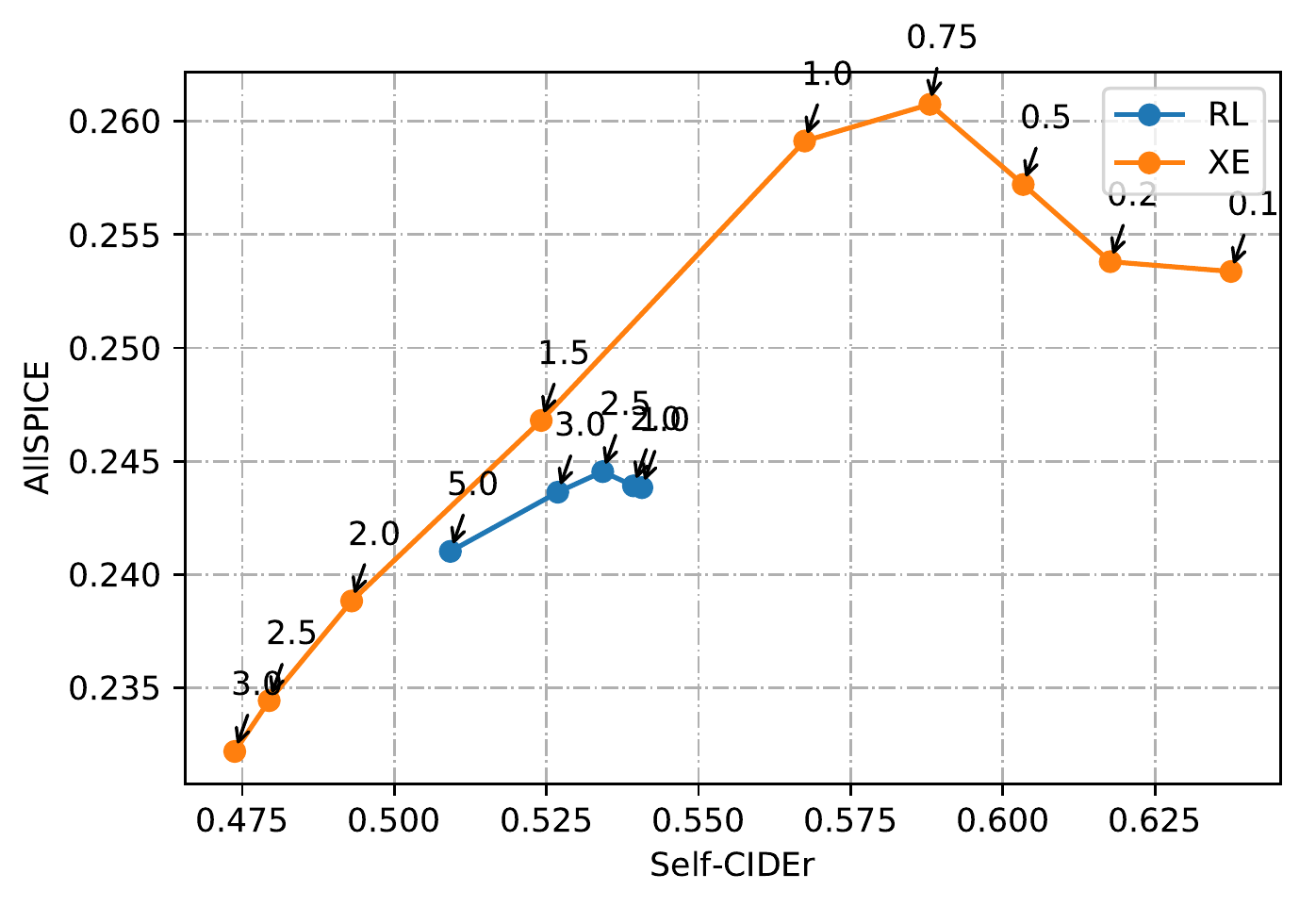}\\
\caption{Oracle CIDEr and AllSPICE of beam search with different temperature, from RL-trained model(blue) and XE-trained model(orange)}
\label{fig:bs}
\end{figure}

Figure \ref{fig:bs} shows how scores change by tuning the temperature for beam search. It's clear that no matter how to tune the BS for RL-trained model, the performance is always lower than that of XE-trained model.

Another interesting finding is unlike other methods, beam search produces more diverse set when having lower temperature. This is because lower temperature makes the new beams to have more chances to be expanded from different old beams instead of having one beam dominating (beams sharing same root).

\section{Diverse beam search}

\begin{figure}[th]
\centering
\includegraphics[height=4cm]{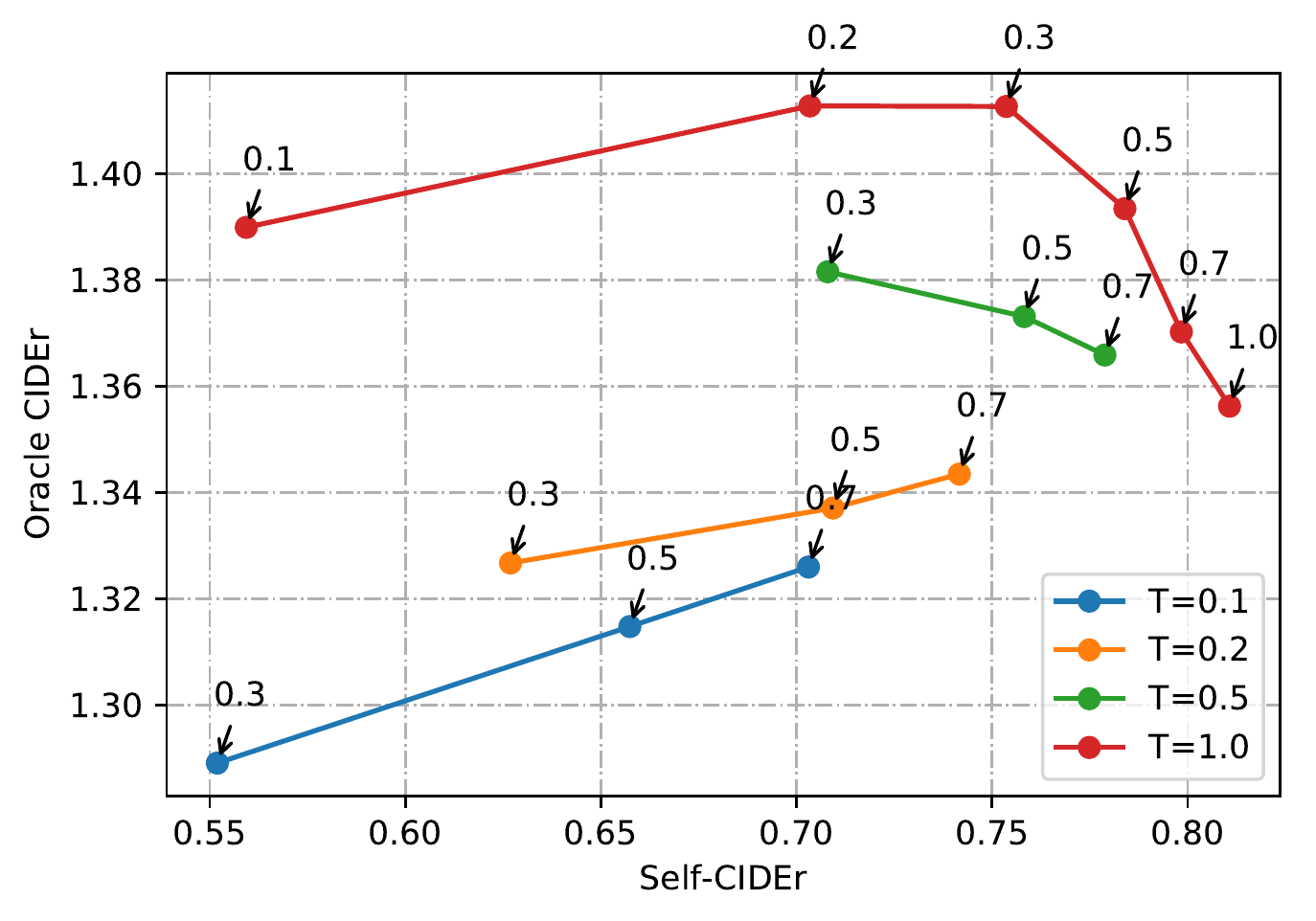}\\
\includegraphics[height=4cm]{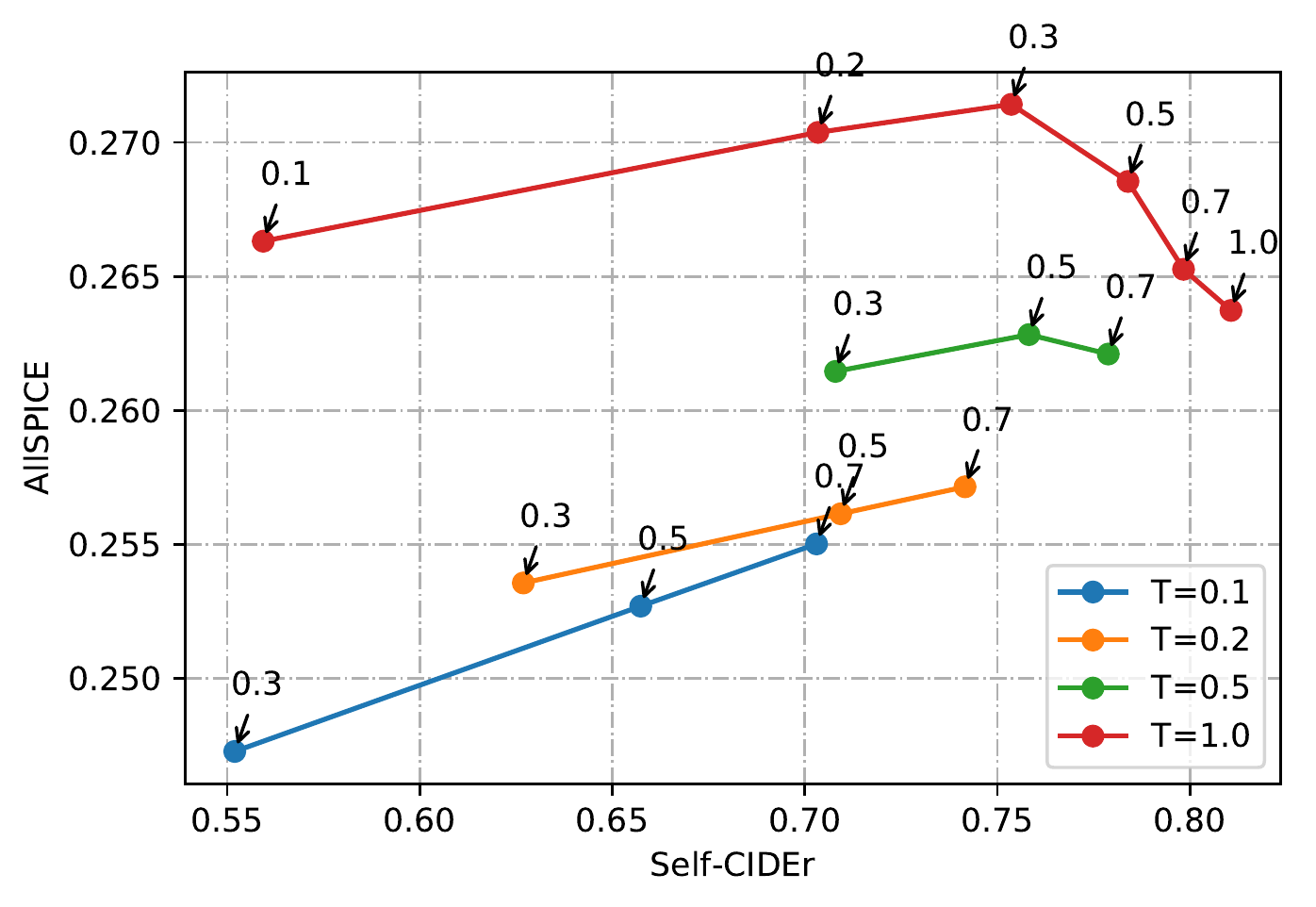}\\
\caption{Diverse beam search. Curve colors correspond to results with
  decoding an XE-trained model with different temperatures; each curve
tracks varying $\lambda$ in DBS.}%
\label{fig:dbs}
\end{figure}

Figure \ref{fig:dbs} shows the performance of oracle CIDEr and AllSPICE with different setting of $\lambda$ and $T$ from a XE-trained model. Figure \ref{fig:dbs_rlxe} compares between XE-trained model and RL-trained model. The RL-trained model is also worse than XE-trained model when using DBS.

\begin{figure}[th]
\centering
\includegraphics[height=4cm]{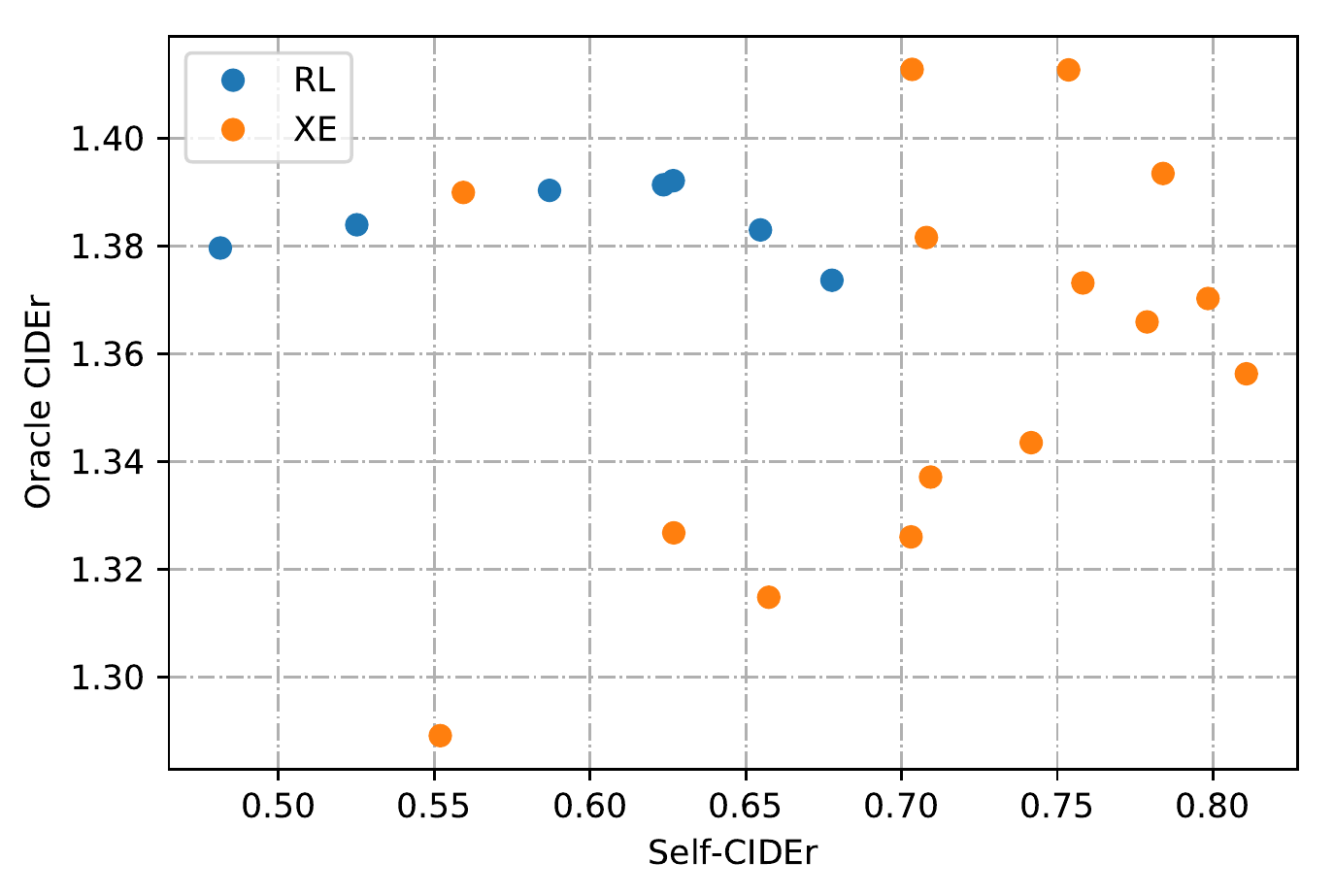}\\
\includegraphics[height=4cm]{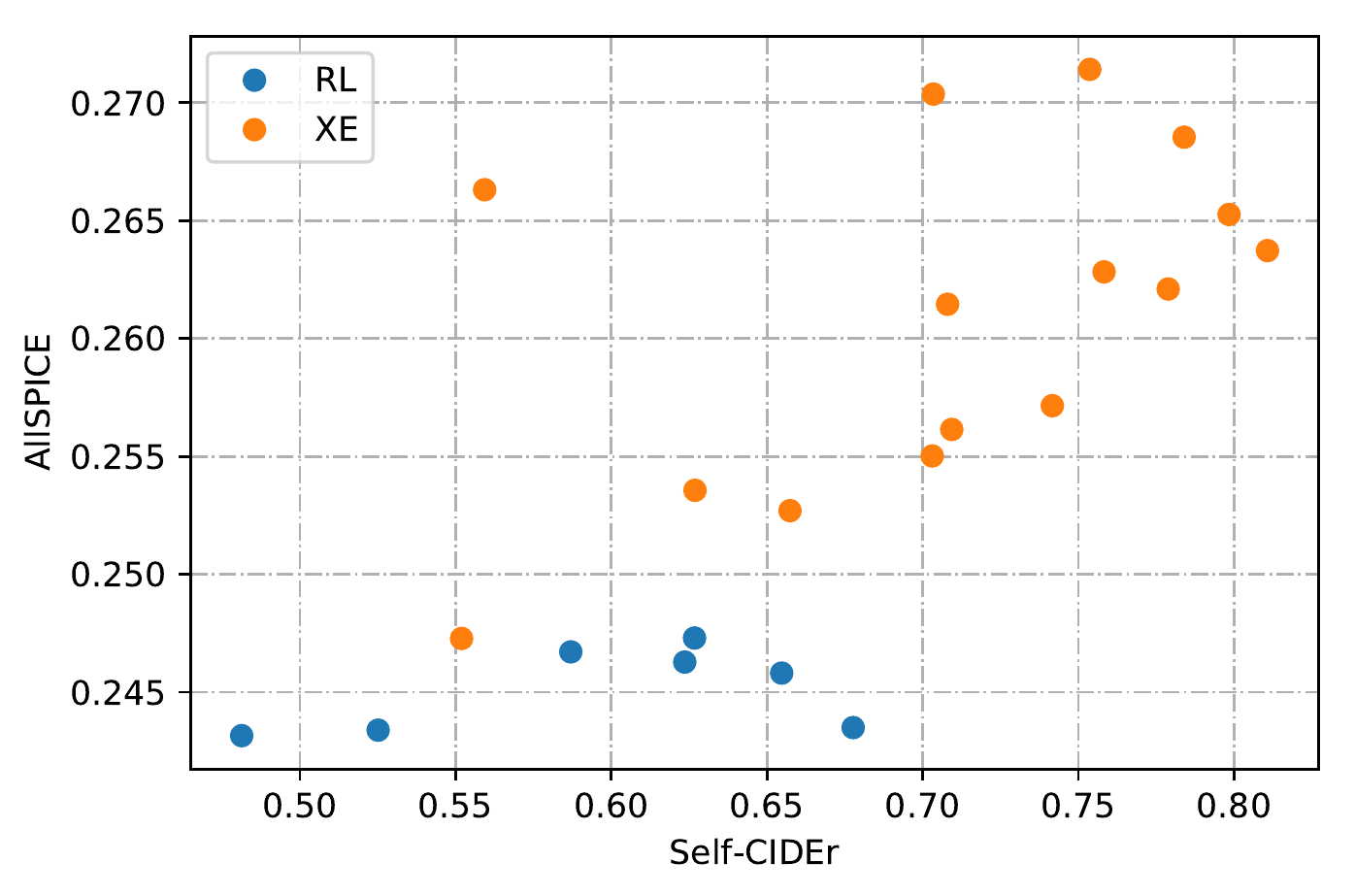}\\
\caption{Diverse beam search result of XE-trained model and RL-trained model. Each dot correspond to a specific $\lambda$ $T$ pair.}%
\label{fig:dbs_rlxe}
\end{figure}

\section{Different models}

Here we compare naive sampling results of different model architectures. We consider 4 different architectures. FC is non-attention LSTM model. ATTN(default in main paper) and ATTN-L are attention-based LSTM with hidden size 512 and 2048. Trans is standard transformer model in \cite{vaswani2017attention, sharma2018conceptual}. They are all trained with XE. 

Table \ref{tab:models} shows the single caption generation result of these 4 architecture: FC $<$ ATTN $<$ ATTN-L $<$ Trans (We include both XE and RL-trained models in the table, but for diversity evaluation we only show results on XE-trained models). Figure \ref{fig:models} shows the naive sampling results for different models. AllSPICE performance on different models has the same order as single caption generation performance.

\begin{table}[htbp]\small
  \centering
    \begin{tabular}{lrrrr}
          & ROUGE & METEOR & CIDEr & SPICE \\
    \midrule
    FC(XE)    & 0.543 & 0.258 & 1.006 & 0.187 \\
    ATTN(XE)  & 0.562 & 0.272 & 1.109 & 0.201 \\
    ATTN-L(XE) & 0.561 & 0.275 & 1.116 & 0.205 \\
    Trans(XE) & 0.563 & 0.278 & 1.131 & 0.208 \\
    \midrule
    FC(RL) & 0.555 & 0.263 & 1.123 & 0.197 \\
    ATTN(RL) & 0.572 & 0.274 & 1.211 & 0.209 \\
    ATTN-L(RL) & 0.584 & 0.285 & 1.267 & 0.219 \\
    Trans(RL) & 0.589 & 0.291 & 1.298 & 0.230 \\
    \bottomrule
    \end{tabular}%
  \caption{The single caption result by different model architectures.}
  \label{tab:models}%
\end{table}%

\begin{figure}[th]
\centering
\includegraphics[height=4cm]{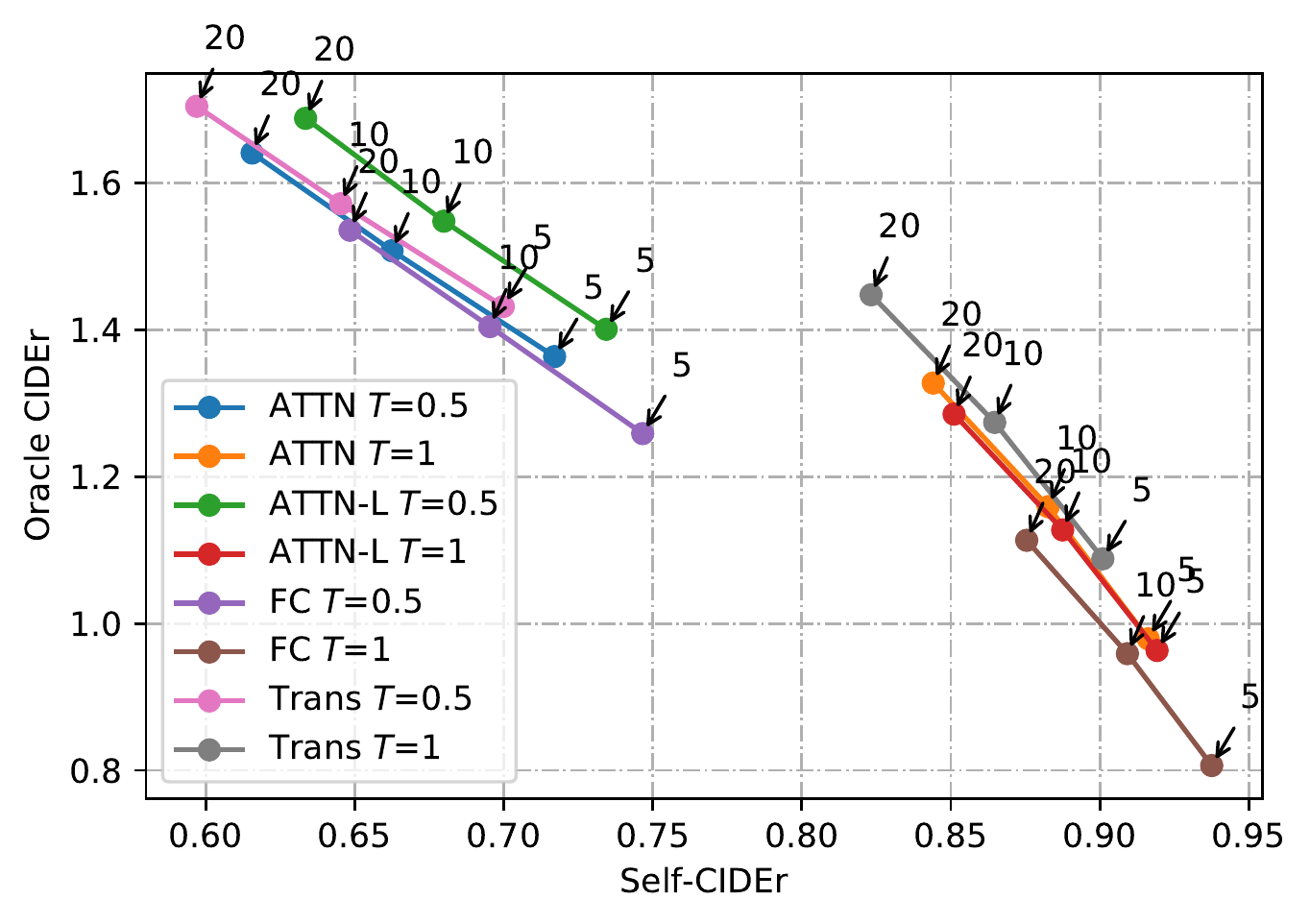}\\
\includegraphics[height=4cm]{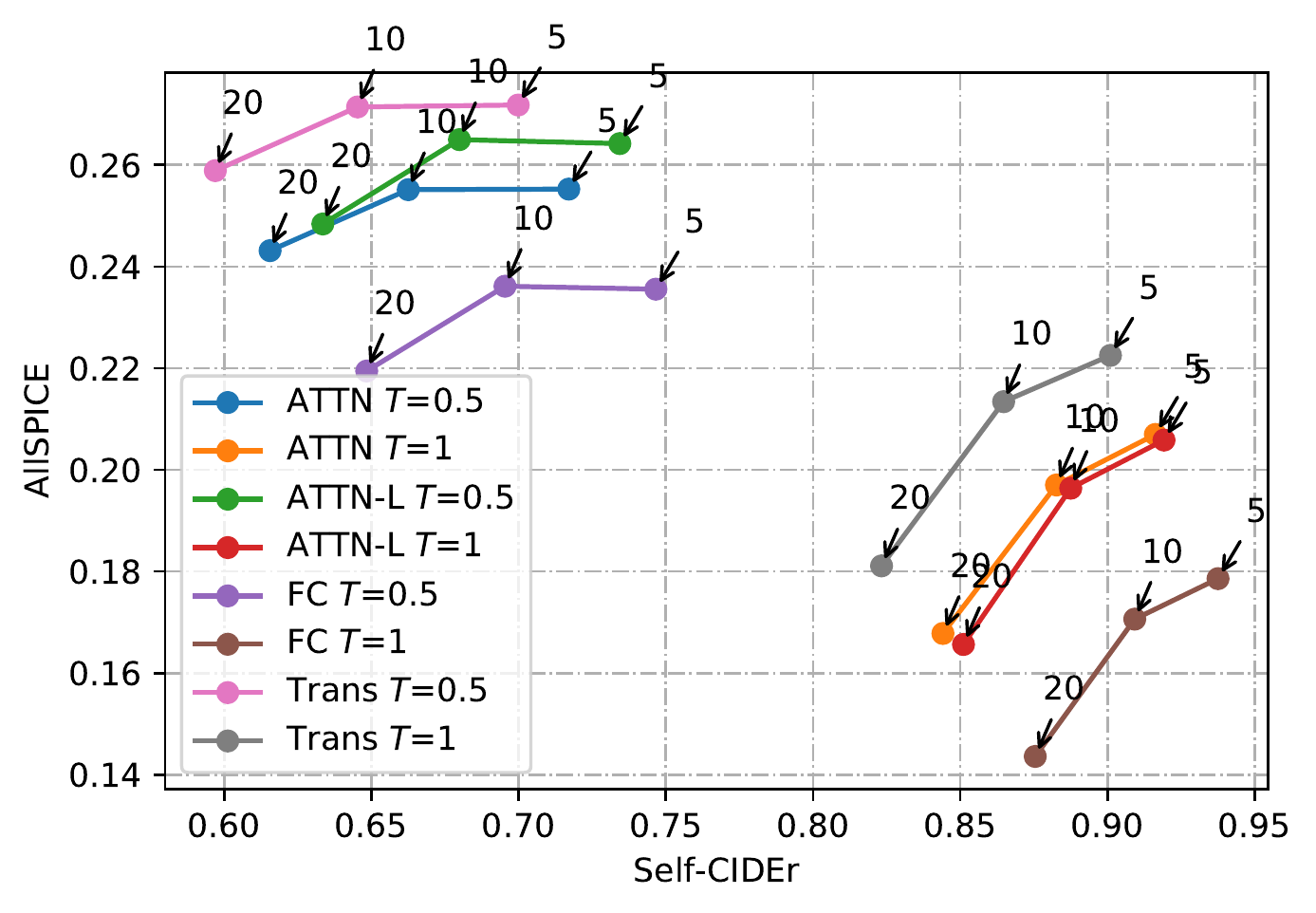}\\
\caption{Oracle CIDEr and AllSPICE of different models. The numbers with arrows indicates the sample size}
\label{fig:models}
\end{figure}

\section{Results on Flickr30k}
Here we also plot similar figures, but on Flickr30k dataset\cite{young-etal-2014-image}. It shows that our conclusions generalize across datasets. This is using the same model as in the main text.

\begin{figure}[th]
\centering
\includegraphics[height=4cm]{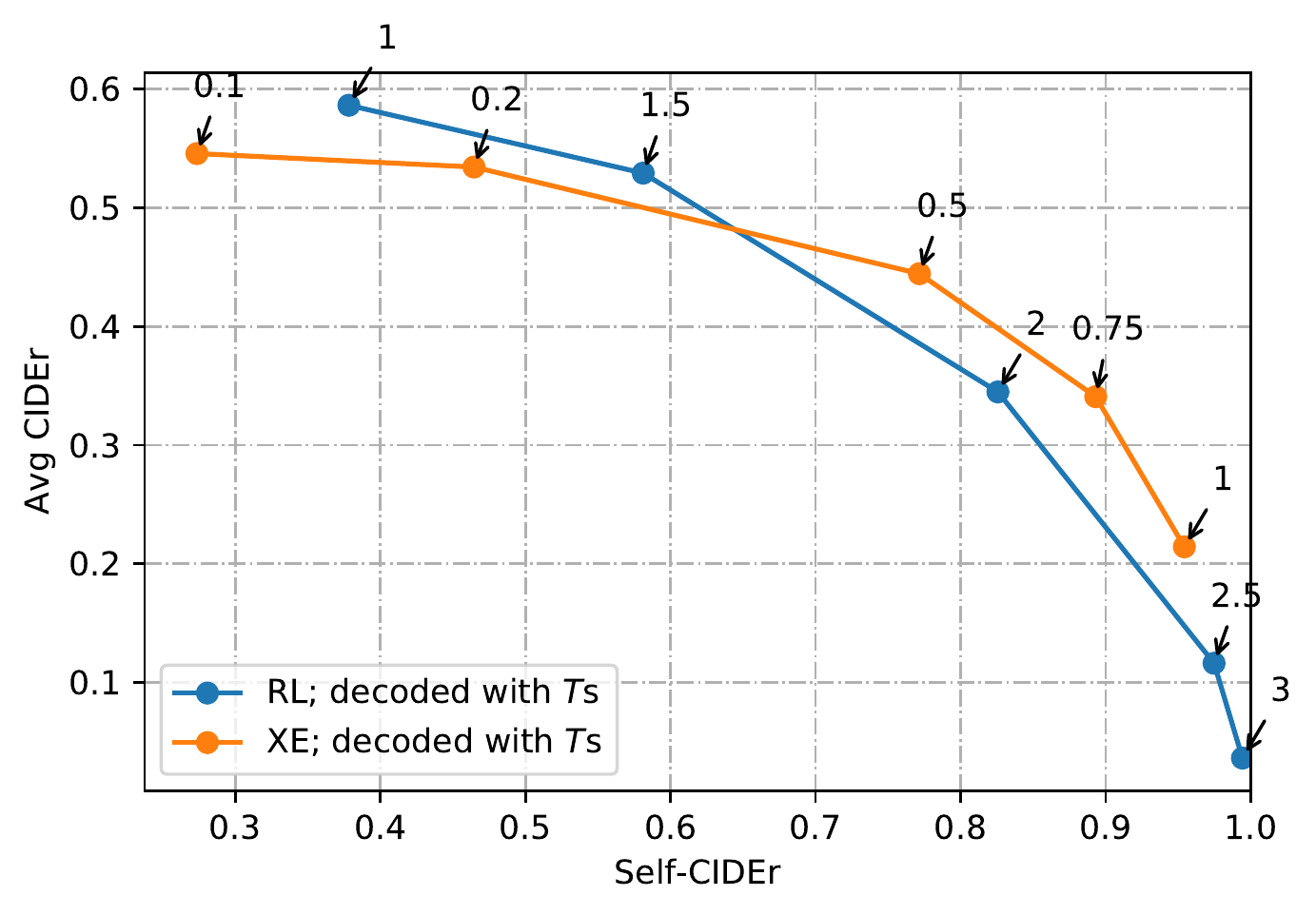}\\
\caption{Diversity-accuracy tradeoff of att2in on Flickr30k. Blue: RL, decoding with different temperatures. 
Orange: XE, decoding with different temperatures.}
\label{fig:f30k1}
\end{figure}

\begin{figure}[th]
\includegraphics[height=4cm]{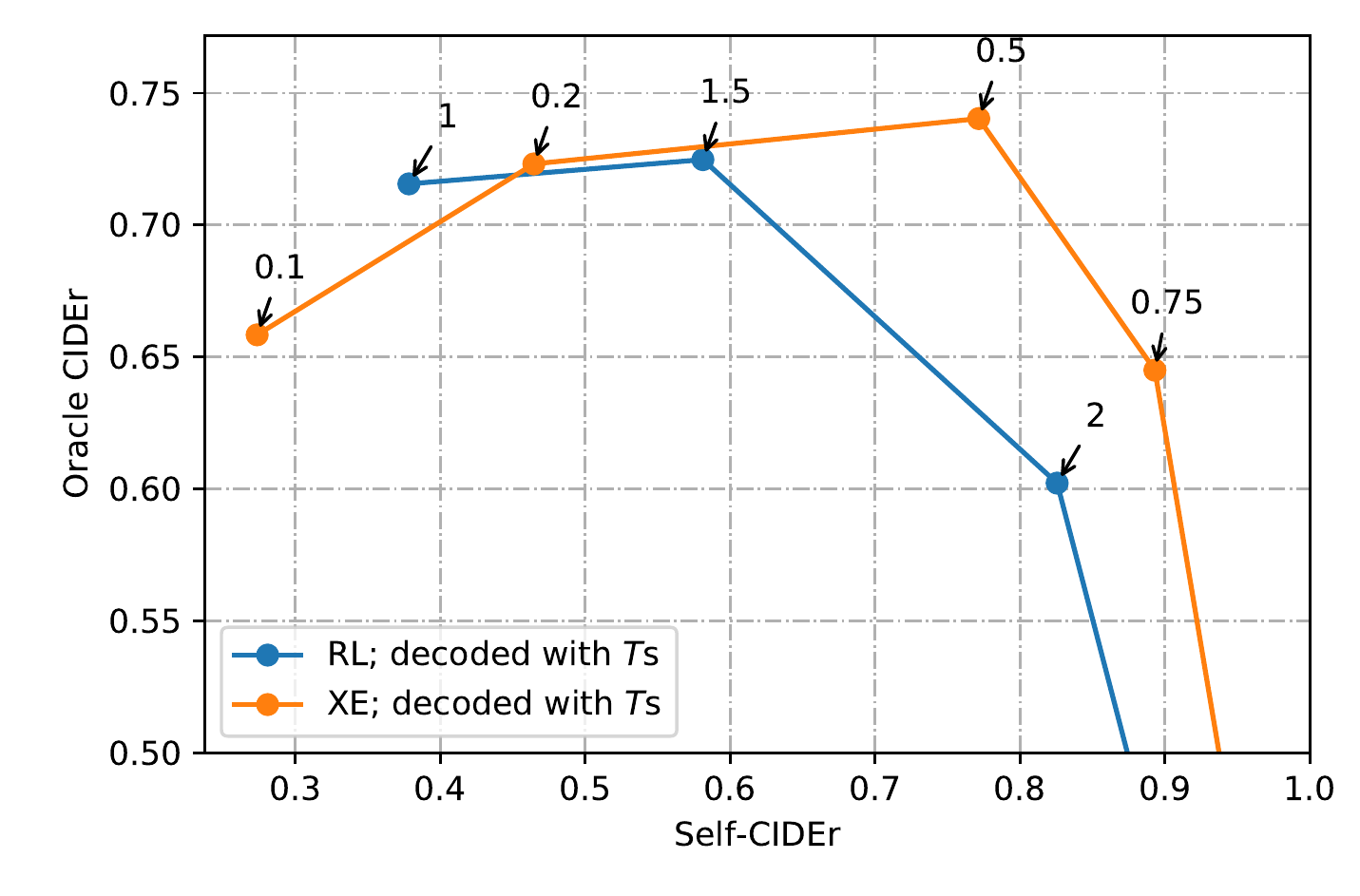}\\
\includegraphics[height=4cm]{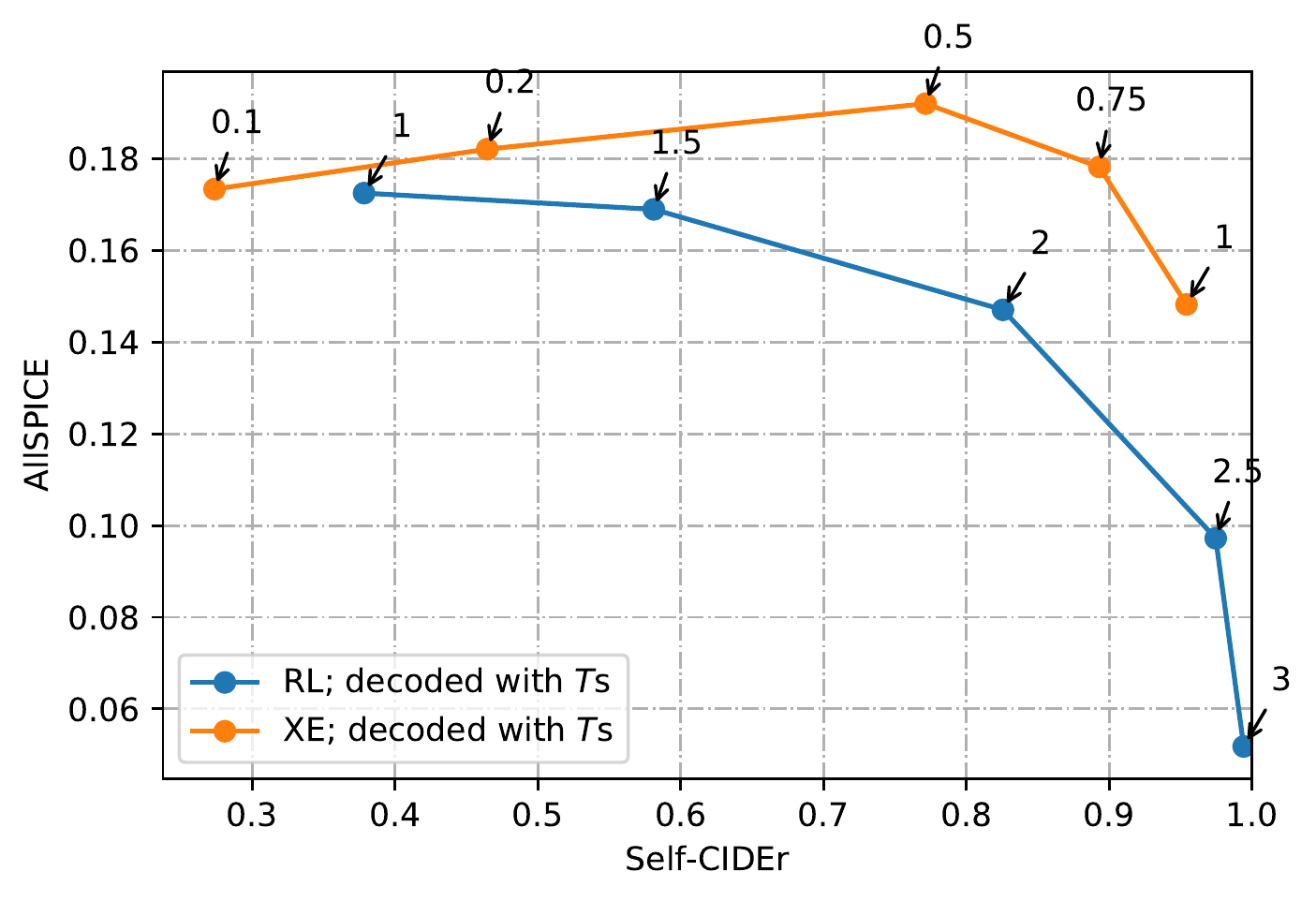}\\
\vspace{-2em}\caption{Oracle CIDEr and \spicen vs. Self-CIDEr  on Flickr30k. See legend for Figure~\ref{fig:f30k1}.}
\label{fig:f30k2}
\end{figure}

\begin{figure}[tb!]
\includegraphics[height=4cm]{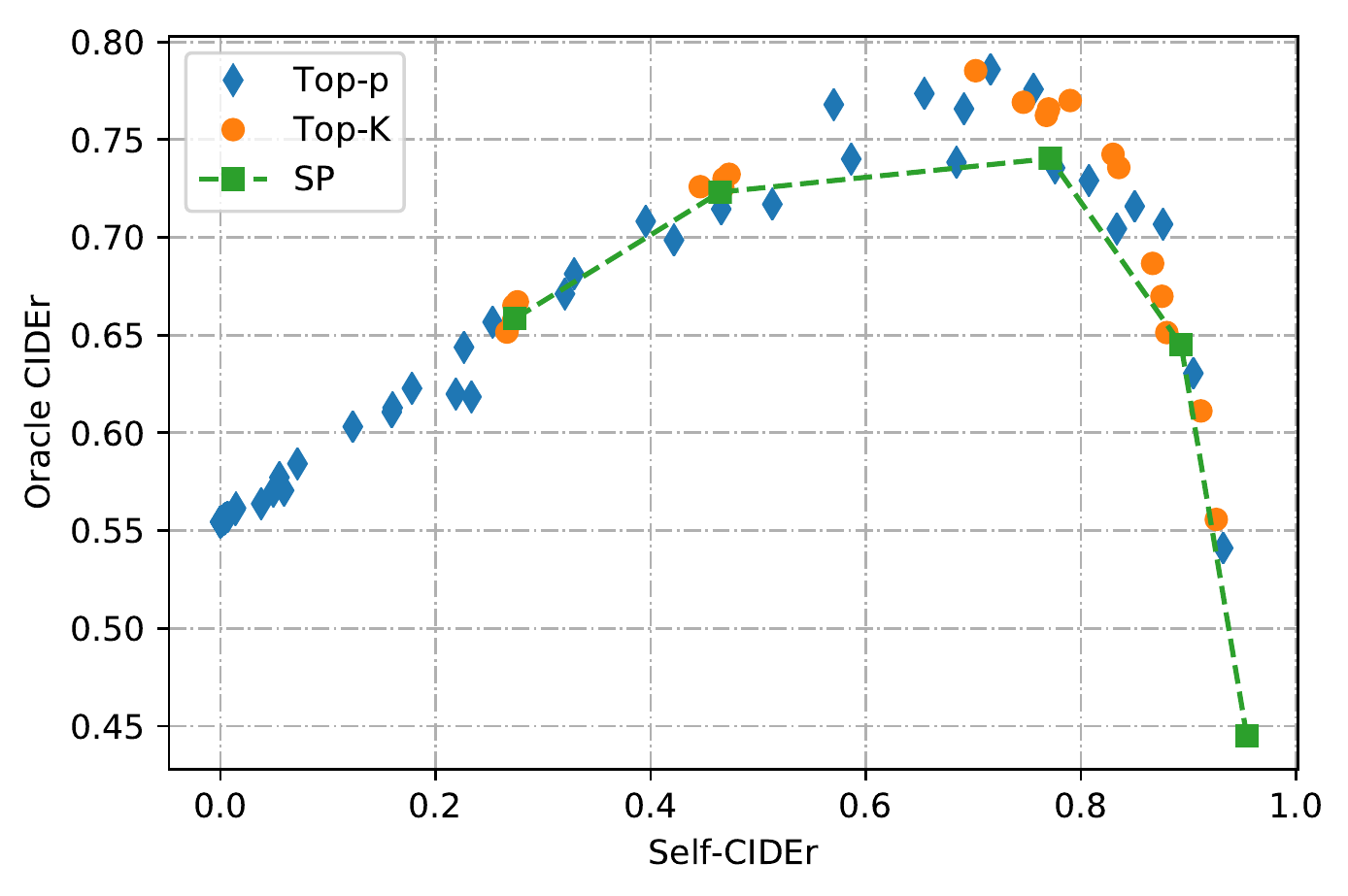}
\hspace{.25em}
\includegraphics[height=4cm]{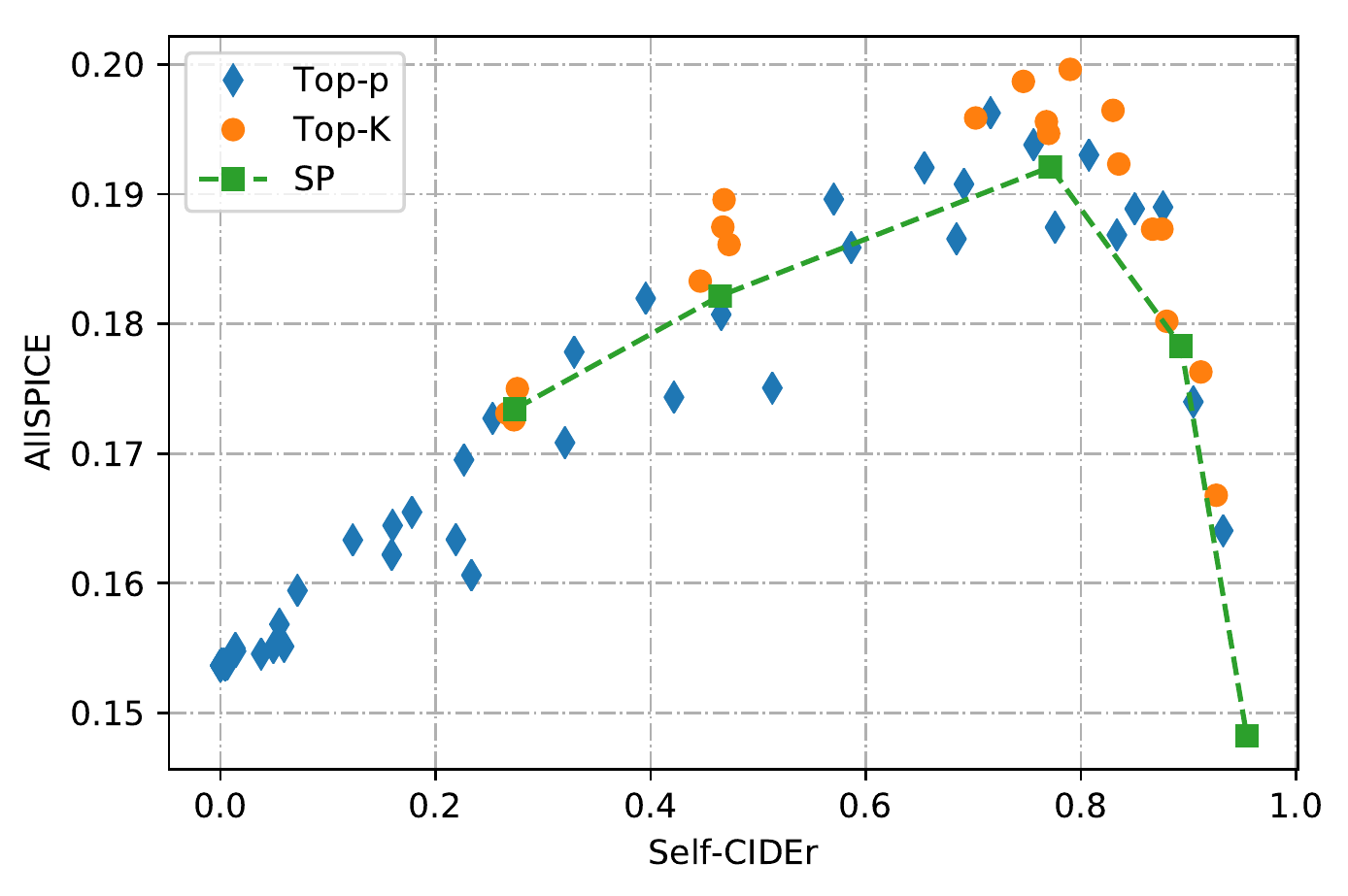}\\
\vspace{-1.5em}\caption{Oracle CIDEr and \spicen of Top-K (orange) and Top-p (blue)
  sampling, compared to naive sampling with varying $T$
  (green) on Flickr30k, for XE-trained model.}%
\label{fig:f30k3}
\end{figure}

\section{Qualitative results}

We show caption sets sampled by different decoding method for 5 different images. The model is attention LSTM model trained with cross entropy loss. 

\begin{figure}
\begin{minipage}[t]{\linewidth}
\setstretch{.75}
\includegraphics[height=4cm]{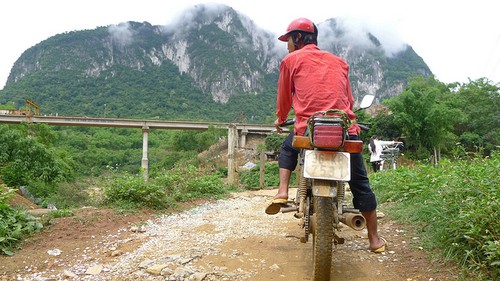}
\end{minipage}
\setstretch{.75}
{\footnotesize
{\bf DBS $\lambda$=3 T=1}:\\
a man riding a motorcycle down a dirt road\\there is a man riding a motorcycle down the road\\man riding a motorcycle down a dirt road\\an image of a man riding a motorcycle\\the person is riding a motorcycle down the road

{\bf BS T=0.75}:\\
a man riding a motorcycle down a dirt road\\a man riding a motorcycle on a dirt road\\a man riding a motorcycle down a rural road\\a man riding a motorcycle down a road\\a man riding a motorcycle down a road next to a mountain

{\bf Top-K K=3 T=0.75}:\\
a man riding a motorcycle down a dirt road\\a person on a motor bike on a road\\a man riding a bike on a path with a mountain in the background\\a man riding a motorcycle down a dirt road with mountains in the background\\a man riding a motorcycle down a road next to a mountain

{\bf Top-p p=0.8 T=0.75}:\\
a man riding a motorcycle down a dirt road\\a person riding a motorcycle down a dirt road\\a man riding a motorcycle down a dirt road\\a man riding a motorcycle down a dirt road\\a man on a motorcycle in the middle of the road

{\bf SP T=0.5}:\\
a man riding a motorcycle down a dirt road\\a man on a motorcycle in the middle of a road\\a person riding a motorcycle down a dirt road\\a man rides a scooter down a dirt road\\a man riding a motorcycle on a dirt road
}
\end{figure}

\begin{figure}
\begin{minipage}[t]{\linewidth}
\setstretch{.75}
\includegraphics[height=4cm]{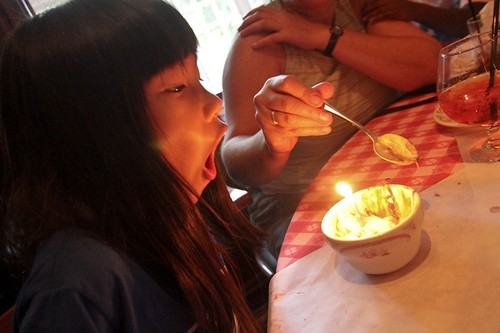}
\end{minipage}
\setstretch{.75}
{\footnotesize
{\bf DBS $\lambda$=3 T=1}:\\
a woman sitting at a table with a plate of food\\two women are sitting at a table with a cake\\the woman is cutting the cake on the table\\a woman cutting a cake with candles on it\\a couple of people that are eating some food

{\bf BS T=0.75}:\\
a woman sitting at a table with a plate of food\\a woman sitting at a table with a cake\\a woman cutting a cake with a candle on it\\a couple of women sitting at a table with a cake\\a couple of women sitting at a table with a plate of food

{\bf Top-K K=3 T=0.75}:\\
a woman sitting at a table with a cake with candles\\a woman cutting into a cake with a candle\\a woman sitting at a table with a fork and a cake\\a woman cutting a birthday cake with a knife\\a woman sitting at a table with a plate of food

{\bf Top-p p=0.8 T=0.75}:\\
two women eating a cake on a table\\a woman cutting a cake with a candle on it\\two women sitting at a table with a plate of food\\a woman cutting a cake with candles on it\\a woman is eating a cake with a fork

{\bf SP T=0.5}:\\
a woman is cutting a piece of cake on a table\\a woman is cutting a cake with a knife\\a woman is eating a piece of cake\\a woman is cutting a cake on a table\\a group of people eat a piece of cake
}
\end{figure}

\begin{figure}
\begin{minipage}[t]{\linewidth}
\setstretch{.75}
\includegraphics[height=6cm]{figures/captions/f3.jpeg}
\end{minipage}
\setstretch{.75}
{\footnotesize
{\bf DBS $\lambda$=3 T=1}:\\
a person riding a bike down a street\\there is a man riding a bike down the street\\the woman is riding her bike on the street\\an image of a person riding a bike\\an older man riding a bike down a street

{\bf BS T=0.75}:\\
a man riding a bike down a street next to a train\\a person riding a bike down a street\\a person riding a bike on a street\\a man riding a bicycle down a street next to a train\\a man riding a bike down a street next to a red train

{\bf Top-K K=3 T=0.75}:\\
a person on a bicycle with a red train\\a man riding a bike down a street next to a red train\\a man riding a bike down a street next to a train\\a man riding a bike down a road next to a train\\a person on a bike is on a road

{\bf Top-p p=0.8 T=0.75}:\\
a woman riding a bicycle in a street\\a woman rides a bicycle with a train on it\\a man on a bicycle and a bike and a train\\a man rides a bike with a train on the back\\a woman is riding her bike through a city

{\bf SP T=0.5}:\\
a woman on a bike with a bicycle on the side of the road\\a person riding a bicycle down a street\\a man on a bicycle near a stop sign\\a man riding a bike next to a train\\a man riding a bike down the street with a bike
}
\end{figure}

\begin{figure}
\begin{minipage}[t]{\linewidth}
\setstretch{.75}
\includegraphics[height=6cm]{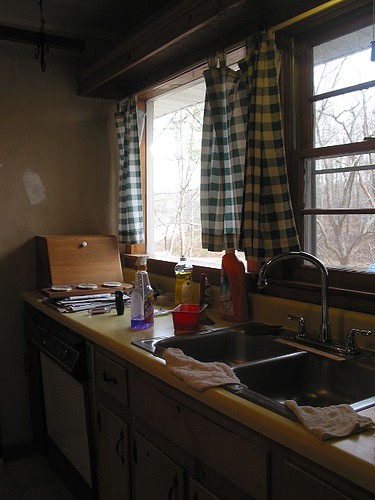}
\end{minipage}
\setstretch{.75}
{\footnotesize
{\bf DBS $\lambda$=3 T=1}:\\
a kitchen with a sink and a window\\there is a sink and a window in the kitchen\\an empty kitchen with a sink and a window\\an image of a kitchen sink and window\\a sink in the middle of a kitchen

{\bf BS T=0.75}:\\
a kitchen with a sink and a window\\a kitchen with a sink a window and a window\\a kitchen sink with a window in it\\a kitchen with a sink and a sink\\a kitchen with a sink and a window in it

{\bf Top-K K=3 T=0.75}:\\
a kitchen with a sink and a mirror\\the kitchen sink has a sink and a window\\a sink and a window in a small kitchen\\a kitchen with a sink a window and a window\\a kitchen with a sink a window and a mirror

{\bf Top-p p=0.8 T=0.75}:\\
a kitchen with a sink a window and a window\\a kitchen with a sink a sink and a window\\a kitchen with a sink and a window\\a kitchen with a sink and a window\\a kitchen with a sink and a window

{\bf SP T=0.5}:\\
a kitchen with a sink a window and a window\\a sink sitting in a kitchen with a window\\a kitchen sink with a window on the side of the counter\\a kitchen with a sink and a window\\a kitchen with a sink and a window
}
\end{figure}

\begin{figure}
\begin{minipage}[t]{\linewidth}
\setstretch{.75}
\includegraphics[height=6cm]{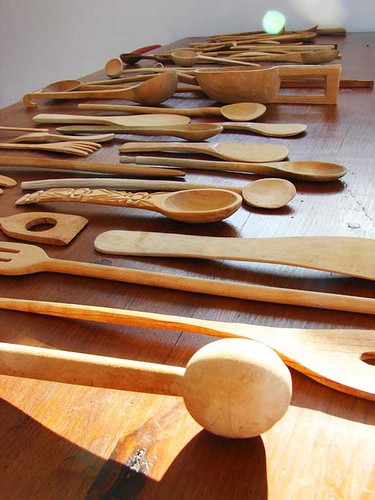}
\end{minipage}
\setstretch{.75}
{\footnotesize
{\bf DBS $\lambda$=3 T=1}:\\
a wooden table topped with lots of wooden boards\\a bunch of different types of food on a cutting board\\there is a wooden cutting board on the table\\some wood boards on a wooden cutting board\\an assortment of vegetables on a wooden cutting board

{\bf BS T=0.75}:\\
a wooden table topped with lots of wooden boards\\a wooden cutting board topped with lots of wooden boards\\a wooden cutting board with a bunch of wooden boards\\a wooden cutting board with a wooden cutting board\\a wooden cutting board with a bunch of wooden boards on it

{\bf Top-K K=3 T=0.75}:\\
a wooden cutting board with a bunch of wooden boards\\a wooden table with several different items\\a wooden cutting board with some wooden boards\\a wooden cutting board with some wooden boards on it\\a bunch of different types of food on a cutting board

{\bf Top-p p=0.8 T=0.75}:\\
a bunch of wooden boards sitting on top of a wooden table\\a wooden cutting board with several pieces of bread\\a wooden cutting board with a bunch of food on it\\a bunch of different types of different colored UNK\\a wooden cutting board with a wooden board on top of it

{\bf SP T=0.5}:\\
a wooden cutting board with knife and cheese\\a wooden table topped with lots of wooden boards\\a wooden cutting board with chopped up and vegetables\\a wooden table topped with lots of wooden boards\\a wooden cutting board with some wooden boards on it
}
\end{figure}

\fi
\end{document}